\documentclass[journal]{IEEEtran}
\usepackage[ruled, linesnumbered]{algorithm2e}
\usepackage{color}
\usepackage{colortbl}
\usepackage{amsmath, amsfonts}
\usepackage{algpseudocode}
\usepackage{array}
\usepackage[caption=false, font=normalsize, labelfont=sf, textfont=sf]{subfig}
\usepackage{textcomp}
\usepackage{stfloats}
\usepackage{url}
\usepackage{verbatim}
\usepackage{graphicx}
\usepackage{amssymb}
\usepackage{graphics} 
\usepackage{epsfig}
\usepackage{booktabs}
\usepackage{multirow}
\usepackage[utf8]{inputenc}
\usepackage{url}
\usepackage{booktabs}
\usepackage{amssymb}
\usepackage{bbding}
\usepackage{pifont}
\usepackage{wasysym}
\usepackage{utfsym}
\usepackage{fontawesome}
\usepackage{booktabs}
\usepackage{multirow}
\usepackage{hyperref}
\usepackage{mathrsfs}
\usepackage[numbers, sort&compress]{natbib}
\usepackage{txfonts}
\begin{document}


\title{Beyond Full Labels: Energy-Double-Guided Single-Point Prompt for Infrared Small \\ Target Label Generation}

\author{Shuai~Yuan,~\IEEEmembership{Student~Member,~IEEE},~Hanlin~Qin,~\IEEEmembership{Member,~IEEE},~Renke Kou,\\~Xiang~Yan,~\IEEEmembership{Member,~IEEE},~Zechaun Li,~Chenxu~Peng,~Huixin~Zhou,~\IEEEmembership{Member,~IEEE}

\thanks{This work was supported in part by the Shaanxi Province Key Research and Development Plan Project under Grant 2022JBGS2-09, in part by the 111 Project under Grant B17035, in part by the Shaanxi Province Science and Technology Plan Project under Grant 2023KXJ-170, in part by the Xian City Science and Technology Plan Project under Grant 21JBGSZ-QCY9-0004, Grant 23ZDCYJSGG0011-2023, Grant 22JBGS-QCY4-0006, and Grant 23GBGS0001, in part by the Aeronautical Science Foundation of China under Grant 20230024081027, in part by the Natural Science Foundation Explore of Zhejiang province under Grant LTGG24F010001, in part by the Natural Science Foundation of Ningbo under Grant 2022J185, in part by the China scholarship council 202306960052, in part by the Technology Area Foundation of China 2021-JJ-1244, 2021-JJ-0471, 2023-JJ-0148. \textit{(Corresponding authors:~Hanlin Qin,~Xiang Yan.)}

Shuai~Yuan, Hanlin~Qin, and Xiang~Yan are with the Xi'an Key Laboratory of Infrared Technology and System, School of Optoelectronic Engineering, Xidian University, Xi'an 710071, China. (email: yuansy@stu.xidian.edu.cn; hlqin@mail.xidian.edu.cn; xyan@xidian.edu.cn)

Renke Kou is with the School of Aviation Engineering, Air Force Engineering University, Xi’an 710038, China. (e-mail: krkoptics@163.com).

Zechuan Li is with the College of Electrical and Information Engineering, Hunan University, Changsha 410082, China (email: lizechuan@hnu.edu.cn).

Chenxu Peng is with the Zhejiang SUPCON Information Co. Ltd., Hangzhou, China (email: chenxupeng@gmail.com).

Huixin Zhou is with the School of Physics, Xidian University, Xi’an 710071,
China (e-mail: hxzhou@mail.xidian.edu.cn).
}
}

\markboth{Journal of \LaTeX\ Class Files,~Vol.~14, No.~8, August~2021}%
{Shell \MakeLowercase{et al.}: A Sample Article Using IEEEtran.cls for IEEE Journals}


\maketitle
\begin{abstract}
We pioneer a learning-based single-point prompt paradigm for infrared small target label generation (IRSTLG) to lobber annotation burdens.
Unlike previous clustering-based methods, our intuition is that point-guided mask generation just requires one more prompt than target detection, i.e., IRSTLG can be treated as an infrared small target detection (IRSTD) with the location hint.
Therefore, we propose an elegant yet effective Energy-Double-Guided Single-point Prompt (EDGSP) framework, aiming to adeptly transform a coarse IRSTD network into a refined label generation method.
Specifically, EDGSP comprises three key modules:
1) target energy initialization (TEI), which establishes a foundational outline to streamline the mapping process for effective shape evolution,
2) double prompt embedding (DPE) for rapidly localizing interesting regions and reinforcing high-resolution individual edges to avoid label adhesion, 
and 3) bounding box-based matching (BBM) for eliminating false masks via considering comprehensive cluster boundary conditions to obtain a reliable output.
In this way, pseudo labels generated by three backbones equipped with our EDGSP achieve 100\% object-level probability of detection (${{P}_{d}}$) and 0\% false-alarm rate (${{F}_{a}}$) on SIRST, NUDT-SIRST, and IRSTD-1k datasets, with a pixel-level intersection over union (IoU) improvement of 13.28\% over state-of-the-art (SOTA) label generation methods.
Further applying our inferred masks to train detection models, EDGSP, for the first time, enables a single-point-generated pseudo mask to surpass the manual labels.
Even with coarse single-point annotations, it still achieves 99.5\% performance of full labeling. Code is available at \url{https://github.com/xdFai/EDGSP}.


\end{abstract}

\section{Introduction}
\label{sec:IN}
\IEEEPARstart{S}{ingle-frame} infrared small target detection (IRSTD) plays a crucial role in maritime rescue, early warning, and high-precision navigation~\cite{MDvsFA},~\cite{Kou_track},~\cite{zhang_IRPruneDet}.
Over the past few years, IRSTD techniques, including convolution neural networks (CNNs)~\cite{Zhang_IS},~\cite{LWIR},~\cite{TMM}, vision transformer (ViT)~\cite{TIPTrans},~\cite{ABC},~\cite{GSTUnet} and vision mamba~\cite{Mamba} have gained significant attention for their powerful abilities in local, global and long-sequence feature modeling.
Among them, most methods rely on segmentation pipelines with pixel-level supervision, highlighting the importance of producing large-scale mask labels~\cite{LELCM},~\cite{IRSTDID},~\cite{LinTIP}.
However, traditional manual pixel-by-pixel annotation is time-consuming and error-prone, primarily due to the limited pixels, low signal-to-clutter ratios, and poorly defined edges of infrared (IR) small targets~\cite{kou_survey},~\cite{zhang_IRSAM},~\cite{TMM2}. 
As a result, reducing the annotation burden for this task has attracted considerable attention.

Recently, single-point labeling has been proposed to reduce the time required for annotating IR small targets by 80\% compared to pixel-level annotation~\cite{MCLC}\footnote{Unless explicitly stated, centroid single-point is used.}.
Surrounding this hot issue, researchers propose two promising solutions.
The first approach directly employs point labels to train a target detection model in a weakly supervised manner. 
For instance, LESPS~\cite{Ying} introduces mapping degradation to extract target contours, leveraging intermediate indicators to direct the evolution of point labels, thereby obtaining detection results with shape representation. 
In contrast, another approach utilizes prompts to guide real-time mask inference independently of the detection model's training~\cite{MCLC},~\cite{MCGC},~\cite{COM}. 
This paradigm allows users to intuitively observe, assess, refine, and further process pseudo masks, making it more commonly preferred.
Typically, Li et al. propose the first infrared small target label generation (IRSTLG) method~\cite{MCLC}, which employs Monte Carlo linear clustering (MCLC) to accumulate results from repeated random noise, progressively reconstructing reliable masks.
Then, multi-scale chain growth clustering (MCGC)~\cite{MCGC} and variational level set~\cite{COM} are further presented to annotate diverse targets in more challenging scenarios.
Though useful, the aforementioned IRSTLG works fall short in providing a systematic evaluation of the obtained pseudo mask, especially concerning false positives and missed annotations.

\begin{figure}[t]
	\centering
	\includegraphics[width=0.475\textwidth]{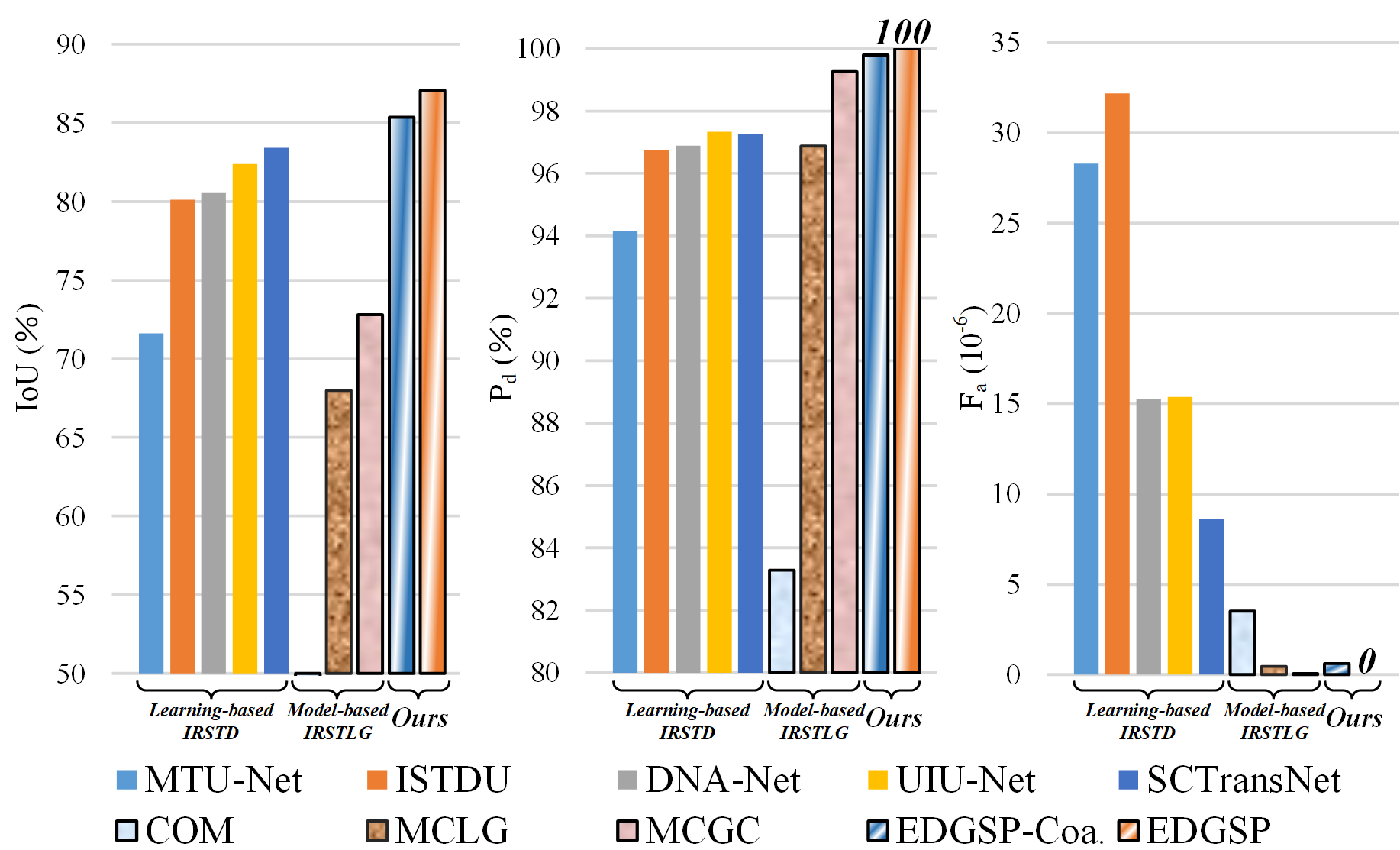}
	\caption{Comparing the result of target detection and label generation on mixed datasets from SIRST, NUDT-SIRST, and IRSTD-1k in $IoU(\%)$, ${{P}_{d}}(\%)$, and ${{F}_{a}}({10}^{-6})$. \textit{EDGSP-Coa.} denotes EDGSP with the coarse single-point prompt.}
	\label{Fig1}
\end{figure}

Since the goal of IRSTLG is to supersede pixel-level manual annotation (ground truth) in training detection model, a natural question arises: \textit{Do these pseudo ground truths truly surpass the conventional masks inferred by detection models?}
To answer this question, we comprehensively evaluate the detection outputs of five IRSTD methods~\cite{MTU},~\cite{ISTDU},~\cite{DNA-Net},~\cite{UIUNet},~\cite{SCTransNet} and the pseudo masks generated by three IRSTLG algorithms~\cite{MCLC},~\cite{MCGC},~\cite{COM} with the same benchmark~\footnote{The detection results and label generation outcomes have an identical format, allowing us to directly evaluate the pseudo masks using target detection metrics.}.
As presented in Fig.~\ref{Fig1}, our experimental results showcase two phenomena.
\begin{itemize}
\itemsep0em
    \item The IOU value of label generation methods is inferior to that of detection models.
    \item Despite knowing the target locations in advance, label generation methods still fail to accurately annotate every target (i.e., object-level ${{P}_{d}} < $  100\%, ${{F}_{a}} > $ 0\%).
\end{itemize}
These findings illustrate that current IRSTLG methods do not meet the fundamental annotation requirements.
As shown in Fig.~\ref{Fig1}, this limitation is largely due to their inability to leverage location cues for deep feature extraction and learning from the already labeled data.
Therefore, we concentrate on establishing a learning-based annotation paradigm for IR small targets in this paper. 
Notably, given that well-designed IRSTD networks have achieved satisfactorily in target shape inference, this work does not build an IRSTLG network from scratch.
Instead, it aims to transform the current IRSTD network into a more advanced label-generation model with the help of single-point prompts, while retaining fully supervised learning.

In fact, point prompts combined with segmentation baselines have been successfully applied to human-in-the-loop object interactive annotation frameworks~\cite{31},~\cite{30},~\cite{interactive}.
The DEXTR~\cite{DEXTR} and IOG~\cite{IOG} employ multiple extreme points and corners to crop the general target region and pair it with the encoded prompt for deep model training.
Recently, the segment anything model (SAM)~\cite{SAM} has demonstrated powerful zero-shot segmentation capabilities, achieved by training on the largest available segmentation dataset.
It encodes multiple types of prompts within a multi-layer transformer-based U-shaped network to map the target and build the long-distance context of the whole image. 
Despite achieving satisfactory results, huge scales and characteristic gaps between regular objectives and IR small targets make these interactive segmentation methods unsuitable for our mission.
We distill these reasons into three areas:
1) Only one pixel within the target area is labeled, which obstacles the cropped-based approach to obtain accurate boundary coordinates~\cite{DEXTR},~\cite{IOG}. Additionally, the lack of a structured target prior representation limits deep models that process entire images in their ability to recover a complete mask~\cite{SAM}.
2) Single-point prompts are always embedded into low-resolution features using cumbersome multi-head attention to align with regions of interest~\cite{HQSAM},~\cite{RobustSAM}. However, IR small targets comprise only 0.15\% of the total image pixels, and their edges are blurred by diffraction effects. Neglecting to model the fine-grained high-resolution contour distinctions making two neighboring targets easily stick and mistakenly labeled as one.
3) Cloud edges, sea phosphorescence, and bright buildings exhibit energy distributions similar to small targets~\cite{Ying2}, current annotation methods lack a stable and well-considered removal component for false annotated masks~\cite{Ying}.



\begin{figure}[t]
	\centering
	\includegraphics[width=0.475\textwidth]{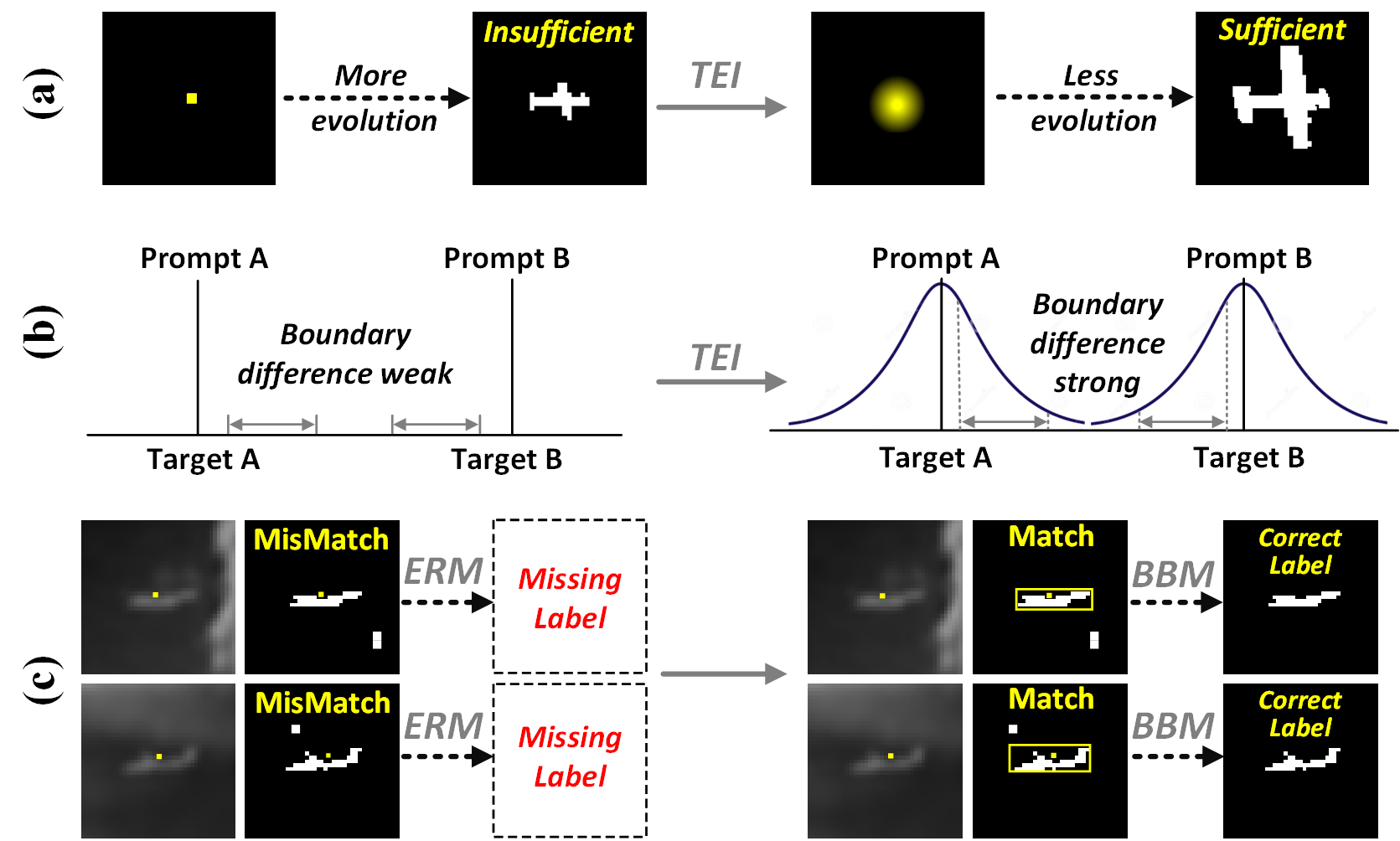}
	\caption{(a) The proposed target energy initialization not only simplifies the shape evolution process but provides a more sufficient target outline representation. (b) Prompts equipped with TEI enhance grayscale and gradient differences between neighboring targets, motivating us to embed the prompt at the model's end to prevent label adhesion. (c) Bounding box-based matching showcases more stable false alarm removal capability than  eight-connective regions matching (ERM)~\cite{Ying}.}
    \vspace{-2mm}
	\label{Fig-1}
\end{figure}


To address these challenges, we propose an energy-double-guided single-point prompt~(EDGSP) framework for IRSTLG.
By integrating target local salience, prompt embedded frequency, and tailored post-processing, EDGSP seamlessly upgrades the basic IRSTD networks into delicate annotation models.
Specifically, our framework consists of three compact components: target energy initialization (TEI) (Section~\ref{1}), double prompt embedding (DPE) (Section~\ref{2}), and bounding box-based matching (BBM) (Section~\ref{3}).
Inspired by the similar texture, energy, and shape distribution of IR small targets~\cite{I1},~\cite{I2},~\cite{I3}, 
TEI expands each point prompt into a Gaussian with local contrast priors, serving as an initial approximation for each target. 
As demonstrated in Fig.~\ref{Fig-1}(a), replacing ``points" with ``spots" not only mitigates the shape evolution gap but also promotes a sufficient target outline formation.
Next, turn our attention to two neighboring targets. 
As presented in Fig.~\ref{Fig-1}(b), the boundary differentiation between adjacent prompts in both grayscale and gradient is enhanced by TEI. 
This observation motivates us to leverage the prompt's edge differences to highlight target independence. 
Consequently, DPE further embeds prompts into the model's final high-resolution mapping layer to prevent shape merging. 
Finally, BBM assesses bounding boxes for each candidate target and matches every point prompt to its respective box by coordinates, aiming to eliminate false masks and obtain the final pseudo mask.
Benefiting from the above structures, our method showcases strong shape prediction capabilities and achieves accurate annotation (${{P}_{d}}$=100\%, ${{F}_{a}}$=0\%) across three datasets - see Fig.~\ref{Fig1}. The main contributions are summarized as follows.


\begin{itemize}
	\itemsep0em

    \item To the best of our knowledge, we present the first study of the learning-based infrared small target annotation paradigm, and introduce EDGSP bridging a crucial link between label generation and target detection task.
 
    \item We propose the TEI to streamline the mapping process for improved shape evolution, DPE to prevent label adhesion by strengthening the separation of neighboring prompts, and BBM to eliminate false masks via comprehensively considering cluster boundary conditions.



	\item For the first time, three backbones equipped with EDGSP achieve accurate annotation on three public datasets.  Meanwhile, the detection task illustrates that IRSTD models trained with our generated label surpass the full label. Even with coarse point annotated, EDGSP achieves 99.5\% detection performance of full label. 
\end{itemize}

\section{RELATED WORK}
\label{Sec2}

\subsection{Infrared Small Target Detection}
IRSTD techniques have been developed over decades, and for a long period, they relied on handcrafted traditional methods such as image filtering~\cite{Tophat},~\cite{MMfilters},~\cite{lvbo} human visual system~\cite{LCM1},~\cite{LCM2},~\cite{LCM3} and low-rank optimization~\cite{lowrank2},~\cite{lowrank1},~\cite{lowrank3},~\cite{Qin}.
Recently, deep learning-based methods have been able to obtain high-level semantics from large amounts of paired data by stacking nonlinear feature extractors, achieving robust detection in complex scenes. 
Since pixel-by-pixel classification is crucial for subsequent target recognition, the segmentation-based IRSTD model is a key emphasis in these data-driven approaches.
Based on a U-shaped structure~\cite{U-Net}, ACM~\cite{ACM} improves the feature fusion strategy of high-level semantics with low-level details using an asymmetric context modulator.
To mitigate the semantic gap between encoders and decoders, DNA-Net~\cite{DNA-Net} adopts dense nested interactive modules to facilitate progressive interaction of cross-scale features and adaptive enhancement of fine-grained features, while UIU-Net~\cite{UIUNet} directly nests the small U-Net into each codec of the large U-shaped structure.
ISNet~\cite{Zhang_IS} proposes a Taylor finite difference-inspired edge block to capture precise outlines of infrared targets by enhancing the shape representation.
To address the limitations in long-range information modeling, MTU-Net~\cite{MTU} employs a self-attention to compute the spatial correlations of fused features, thereby constructing a global representation of the image.
Most recently, SCTransNet~\cite{SCTransNet} explicitly interacts and enhances full-level encoder features by spatial-channel cross transformer blocks to effectively suppress interference.

In summary, existing research emphasizes designing tight structures to achieve high performance~\cite{Zhangtianfang}. However, the labor-intensive task of pixel-level annotation for infrared small targets hampers the advancement of these data-dependent methods.
Therefore, this work concentrates on alleviating the burden of infrared small target labeling, i.e., using the single-point prompt to generate the target's pseudo mask.

\subsection{Infrared Small Target Label Generation}
To the best of our knowledge, there are only three IRSTLG efforts for directly annotating IR small targets.
These efforts are all model-based single-point prompt methods.
As a pioneer, MCLC~\cite{MCLC} first employs a Monte Carlo linear clustering process with random noise and averages the clustering results to obtain reliable pseudo labels. 
To enhance annotation robustness, MCGC~\cite{MCGC} adaptive recovers small targets with multi-scale chain growth clustering under the randomly given single-point labels, which can be adapted to different numbers, scales, shapes, and intensities of targets. 
Subsequently, Li et al.~\cite{COM} design a variational level set with an expected difference energy function to address the excessive regularization causing target contours to disappear.

Unlike the works mentioned above, we present a learning-based  IRSTLG paradigm that harnesses the contour inference strengths of existing IRSTD models, artfully transforming the target detection network into a high-level label generation method through the use of single-point prompts.


\begin{figure*}[t!]
    \centering
    \includegraphics[width=0.99\textwidth]{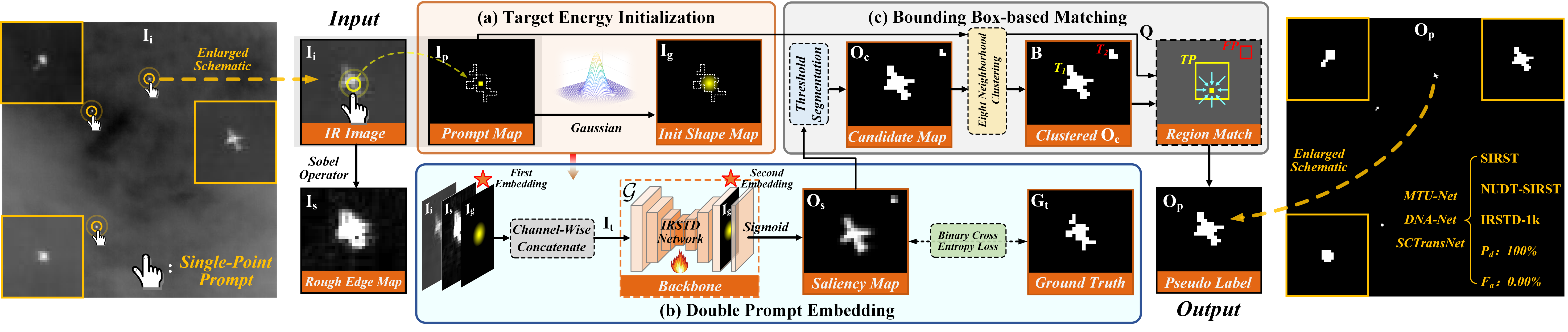}
    \caption{Illustration of the proposed energy-double-guided single-point prompt (EDGSP) for infrared small target label generation. It consists of three parts. (a) Target energy initialization (TEI) for sufficient shape evolution, (b) double prompt embedding (DPE) to reinforce individual differences and prevent mask adhesion, and (c) bounding box-based matching (BBM) to eliminate false annotations.}
    \label{Fig2}
\end{figure*}

\subsection{Point Prompt Interactive Segmentation on Image Processing}
\label{IIC}
The technique most similar to ours is point prompt interactive segmentation~\cite{GrabCut},~\cite{xu2016deep},~\cite{interactive}.
For example, DEXTR~\cite{DEXTR} uses four extreme points for segmentation, the interest region is cropped and the Gaussian heatmap is utilized as the encoding form of prompts.
Similarly, IOG~\cite{IOG} uses the approximate center of the target as the inside point and a set of symmetrical corner points as outside points. After cropping the target of interest, the image, along with the heatmap is fed into the segmentation model.
Luo et al.~\cite{MIDeepSeg} uses the exponentiated geodesic distance transform to convert medical images into Euclidean distance maps for improved foreground-background separation.
However, the methods mentioned above typically require multiple points to precisely locate target edges and crop the regions of interest. 
In contrast, our single-point annotation approach directly processes the entire image without cumbersome cropping.
Recently, a series of segment anything model (SAM)~\cite{SAM},~\cite{MedSAM},~\cite{RobustSAM} has gained recognition for its flexibility and generalization capabilities. 
SAM is a transformer-based model designed for zero-shot semantic segmentation, utilizing various types of prompts to encode key areas.
For single-point prompts, SAM typically encodes prompt coordinates and performs complex interactions with global image features across multiple transformer layers, aiming to better align the mapped area of the prompt with the target region~\cite{HQSAM}.


Despite the success of the above work in generic target segmentation, the unique characteristics of infrared small targets render these approaches ineffective.
Different from them, EDGSP is the first deep annotation model tailored for infrared small targets.
Our approach introduces TEI and DPE to address insufficient single-target contour evolution and neighbor-target differentiation, and utilize a box-based matching approach to eliminate false masks without bells and whistles.



\section{Proposed Method}
Our objective is to develop a mask annotation framework for the infrared small target with a single-point guide.
To achieve this, we present the energy-double-guided
dingle-point prompt framework to transform the IRSTD network into a more advanced label-generation model - Fig.~\ref{Fig2}.
Our method accepts two inputs: an entire infrared image ${\mathbf{{I}_{i}}} \in \mathbb{R}^{{1} \times {H} \times {W}}$ that requires annotation and a coordinate $(x_{0},y_{0})$ obtained by user-specified click within the target region.
The proposed target energy initialization (TEI) is first utilized to automatically generate a coarse outline of the target to streamline the mapping process for improved shape evolution. In the double prompt embedding (DPE) module, the encoded prompt is embedded into the IRSTD backbone twice to localize interesting regions and reinforce the high-resolution individual edges to avoid label adhesion. During the model inference, the bonding box-based matching (BBM) function is utilized to eliminate false masks and obtain the final pseudo mask, which is used as the ground truth for training the detection model. The following sections provide a detailed explanation of each module.


\subsection{Target Energy Initialization}
\label{1}
The long-range imaging mechanism in infrared systems introduces significant diffraction effects, causing small targets to exhibit a nearly isotropic Gaussian distribution~\cite{Mian}. 
Leveraging this physical phenomenon alongside the local contrast prior~\cite{LCM2}, we predefine a basic profile for each target to simplify the original point-to-mask mapping.
As shown in Fig.~\ref{Fig2}, given an input IR image ${\mathbf{{I}_{i}}} \in \mathbb{R}^{{1} \times {H} \times {W}}$ and the coordinate $(x_{0},y_{0})$ of the user's click, the prompt map ${\mathbf{{I}_{p}}} \in \mathbb{R}^{{1} \times {H} \times {W}}$ is defined as:
\begin{equation}
{\mathbf{{I}_{p}}}(x, y) = 
\begin{cases}
 1, & \text{if}~(x, y) = (x_{0},y_{0})  \\
 0, & \text{otherwise}
\end{cases}
\end{equation}
in which \( x \in [1, H] \) and \( y \in [1, W] \)\footnote{Since the single-point prompt is a pixel within the target, it can also be derived from the already labeled ground truth ${\mathbf{{G}_{t}}} \in \mathbb{R}^{{1} \times {H} \times {W}}$.}. 
Next, we define a two-dimensional Gaussian function $G(x, y)$ centered at a given coordinate $(x_0, y_0)$, with the following expression:
\begin{equation}
   G(x, y)\big|_{(x_0, y_0)} = \exp\left(-\frac{(x - x_0)^2 + (y - y_0)^2}{2\sigma^2}\right)
\end{equation}
where $\sigma$ denotes the peak half-width of the Gaussian spread. Then, the init shape map ${\mathbf{{I}_{g}}} \in \mathbb{R}^{{1} \times {H} \times {W}}$ that containing all Gaussian spots can be represented as:
\begin{equation}
{\mathbf{{I}_{g}}}(x, y) = \sum_{(x_0, y_0) \in S}G(x, y)\big|_{(x_0, y_0)}
\end{equation}
where \( S = \{(x, y) \mid {\mathbf{{I}_{p}}}(x, y) = 1\} \) denotes the set of all points in \( I_p \) with a pixel value of 1.
This expansion shifts from shaping area targets with points to refining contours using Gaussian blobs, aligning the prior distribution of the object, and aiding the model in learning complete shape representations. 
In Section~\ref{TEI}, we will discuss the different forms of prompt extensions in detail.

\subsection{Double Prompt Embedding}
\label{2}
In generic object single-point prompt works, the prompt always interacts with global image features across transformer layers after position and label encoding. The multi-head attention is utilized to understand the tokens related to the point prompt and map them to the same region~\cite{SAM},~\cite{HQSAM},~\cite{RobustSAM}. 
However, due to the limited pixels, the prompts and IR small targets are easily spatially aligned. In this case, redundant embeddings instead consume computational resources and reduce scalability. 
Therefore, we adopt the channel-wise concatenation as a simple but effective embedded form to align semantics.
Without loss of generality, given a U-shaped IRSTD network $\mathcal{G}$, encoded prompt ${\mathbf{{I}_{g}}}$ embed into $\mathcal{G}$ twice, as shown in Fig.~\ref{Fig2}(b).
Considering that edge information can guide shape construction~\cite{Zhang_IS}, the Sobel operator $\mathcal{S}$ is first applied to the original infrared image ${\mathbf{{I}_{i}}}$ to obtain a rough edge map ${\mathbf{{I}_{s}}}$. 
Next, ${\mathbf{{I}_{i}}}$, ${\mathbf{{I}_{s}}}$, and the init shape map ${\mathbf{{I}_{g}}}$ are concatenated among the channel dimension as a three-channel input image ${\mathbf{{I}_{t}}} \in \mathbb{R}^{{3} \times {H} \times {W}}$:
\begin{equation}
{\mathbf{{I}_{t}}}= Concat({\mathbf{{I}_{i}}},\mathcal{S}({\mathbf{{I}_{i}}}),{\mathbf{{I}_{g}}})
\end{equation}
Then, the penultimate layer features ${\mathbf{L}}$ of $\mathcal{G}$  are obtained as follows:
\begin{equation}
{\mathbf{L}}= {\mathcal{G}^{Conv}_{w/o}}({\mathbf{{I}_{t}}})
\end{equation}
where ${\mathcal{G}^{Conv}_{w/o}}$ stands for $\mathcal{G}$ that without the last convolution layer. 
Since the grayscale and gradient differentiation between adjacent prompts in ${\mathbf{{I}_{g}}}$ already enhanced by TEI,
${\mathbf{{I}_{g}}}$ is utilized as a semantic map that emphasizes the target quantity and edge in the second embedding. It is concatenated with the ${\mathbf{L}}$ along the channel dimension to refine the model’s final mapping and prevent close-range target sticking. 
\begin{equation}
{\mathbf{{O}_{s}}}= {Sigmoid}({Conv}(Concat({\mathbf{L}}, {\mathbf{{I}_{g}}})))
\end{equation}
where ${\mathbf{{O}_{s}}} \in \mathbb{R}^{{1} \times {H} \times {W}}$ represents the saliency map, ${Conv}(\cdot)$ denotes the last layer of $\mathcal{G}$, typically constructed with a $1\times1$ or $3\times3$ convolution that squeezes channel and preserves the spatial dimensions. 
The binary cross entropy loss is then employed to compute pixel-wise differences between the ${\mathbf{{O}_{s}}}$ and the ground truth ${\mathbf{{G}_{t}}}$.


\begin{algorithm}[t]
\caption{Bounding Box Matching (BBM)}
\label{alg1}
\KwData{Clustered candidate map \( \mathbf{B} \in \mathbb{R}^{1 \times H \times W} \) containing clusters \( \{ T_{1}, T_{2}, \ldots , T_{m} \} \); Clustered prompt map \( \mathbf{Q} \in \mathbb{R}^{1 \times H \times W} \) containing clusters \( \{ G_{1}, G_{2}, \ldots , G_{n} \} \).}
\KwResult{Pseudo label \( \mathbf{{O}_{p}} \in \mathbb{R}^{1 \times H \times W} \).}

\textbf{Initialize} \( \mathbf{U} = \mathbf{B} \)\\
\For{$k \gets 1$ \textbf{to} $n$}{
    \( (x_{0}, y_{0}) = \text{GetCenter}(G_{k}) \);
    
    \textcolor{olive}{\tcp{Get center coordinates of cluster \( G_{k} \)}}
    \For{$j \gets 1$ \textbf{to} $m$}{
        \( (x_{1}, y_{1}, x_{2}, y_{2}) = \text{GetBoundingBox}(T_{j}) \);
        
        \textcolor{olive}{\tcp{Get bounding box coordinates of cluster \( T_{j} \)}}
        \If{\( x_{1} < x_{0} < x_{2} \) \textbf{and} \( y_{1} < y_{0} < y_{2} \)}{
            \( \mathbf{U} = \mathbf{U} - T_{j} \); 
            
            \textcolor{olive}{\tcp{Remove matched region from \( \mathbf{U} \)}}
            \textbf{Break}
        }
    }
}
\( \mathbf{{O}_{p}} = \mathbf{B} - \mathbf{U} \)
\end{algorithm}

\subsection{Bounding Box-based Matching}
\label{3}
In earlier research, eight connective regions matching is commonly used to exclude false alarms during the label evolution process~\cite{Ying}, ensuring that evolution information consistently applies to the initial point.
However, this strategy ignores the cases where the point prompt is located on or outside the target boundary, leading to incorrect processing results (detailed in Fig.~\ref{Fig-1}(c)).
We tackle this issue by adding a compact design.
As shown in Fig.~\ref{Fig2}(c), given the saliency map ${\mathbf{{O}_{s}}} \in \mathbb{R}^{{1} \times {H} \times {W}}$ which is processed by the label generation model and the prompt map ${\mathbf{{I}_{p}}} \in \mathbb{R}^{{1} \times {H} \times {W}}$, we first segment ${\mathbf{{O}_{s}}}$ to obtain the candidate map ${\mathbf{{O}_{c}}} \in \mathbb{R}^{{1} \times {H} \times {W}}$, it is defined as:
\begin{equation}
{\mathbf{{O}_{c}}}= {\mathbf{{A}_{m \times n}}}\odot({\mathbf{{O}_{s}}}>{{{\tau}_{S}}})
\end{equation}
where $\odot$ represents element-wise multiplication, ${\tau}_{S}$ is the threshold of the segmentation, and ${\mathbf{{A}_{m \times n}}} \in \mathbb{R}^{{1} \times {H} \times {W}}$ is an all-ones matrix with the same dimensions of ${\mathbf{{O}_{s}}}$. Next, we apply  the eight-neighborhood clustering algorithm $\mathcal{E}$ on ${\mathbf{{I}_{g}}}$ and ${\mathbf{{O}_{c}}}$ as follows:
\begin{equation}
{\mathbf{B}}=\{{{T}_{1}, {T}_{2}, \ldots , {T}_{m}}\}= \mathcal{E}({\mathbf{{O}_{c}}})
\end{equation}
\begin{equation}
{\mathbf{Q}}=\{{{G}_{1}, {G}_{2}, \ldots , {G}_{n}}\}= \mathcal{E}({\mathbf{{I}_{g}}})
\end{equation}
where the ${\mathbf{B}} \in \mathbb{R}^{{1} \times {H} \times {W}}$ and ${\mathbf{Q}} \in \mathbb{R}^{{1} \times {H} \times {W}}$ denote the clustered ${\mathbf{{O}_{c}}}$ and ${\mathbf{{I}_{g}}}$, $\{{{T}_{1}, {T}_{2}, \ldots , {T}_{m}}\}$ and $ \{{{G}_{1}, {G}_{2}, \ldots , {G}_{n}}\}$ represent the set of clustered regions in ${\mathbf{B}}$ and ${\mathbf{Q}}$, separately.
Then, the bounding box-based matching algorithm is used to iteratively match each cluster in ${\mathbf{Q}}$ and ${\mathbf{B}}$,  based on bounding box overlap in \textbf{Algorithm}~\ref{alg1}.
Among them, $GetCenter(\cdot)$ and $GetBoundingBox(\cdot)$ are functions that retrieve the center coordinates and bounding box coordinates of clusters, respectively.
By removing regions of ${\mathbf{B}}$ that overlap with ${\mathbf{Q}}$, we obtain the final annotated mask ${\mathbf{{O}_{p}}}$.
Section~\ref{BBMABL} comprehensively discusses three types of false alarm elimination mechanisms in both coarse and centroid modes.



\section{Experiments}
In this section, we first introduce the evaluation metrics and experimental details.
Then, we compare the proposed EDGSP with several state-of-the-art point prompt label generation methods. 
Next, the inferred pseudo masks are used to train IRSTD models, providing an indirect assessment of the practical detection value of the generated pseudo labels~\cite{MCLC},~\cite{MCGC},~\cite{COM}.
Simultaneously, Section~\ref{Detection} demonstrates how EDGSP consistently enhances performance when transitioning from detection models to annotation models. 
Following this, we perform a comprehensive analysis and ablation studies to investigate the effective strategies of our model in Section~\ref{Cropping} and Section~\ref{ABL}. Finally, we discuss the method's efficiency.


\begin{figure}[t]
    \centering
    \includegraphics[width=0.49\textwidth]{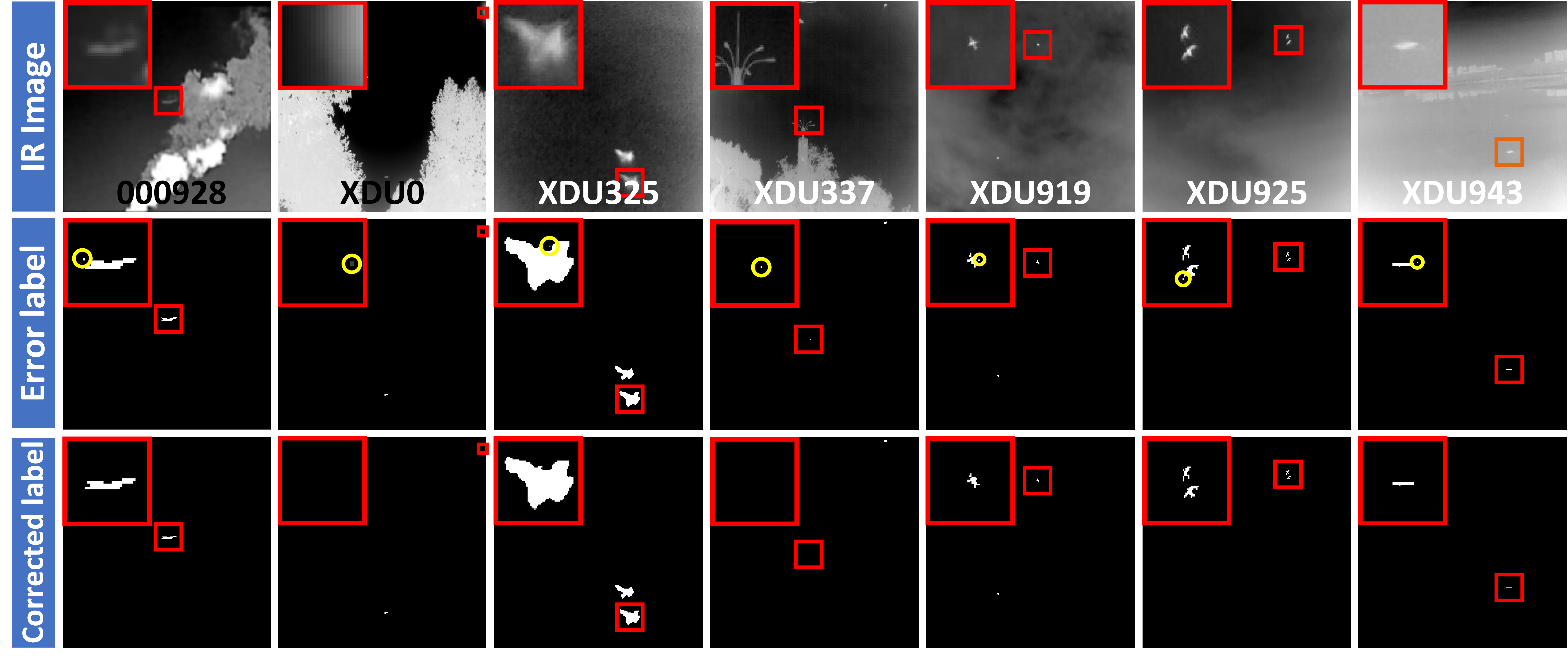}
    \caption{Visualization and correction of seven errors in the NUDT-SIRST~\cite{DNA-Net} and IRSTD-1k~\cite{Zhang_IS} datasets. Error annotations are highlighted in yellow circles.}
    \label{Fig4}
\end{figure}

\begin{table*}[t]
\renewcommand{\arraystretch}{1.3}
\caption{Comparison with SOTA IRSTLG methods on mixed SIRST, NUDT-SIRST, and IRSTD-1k in terms of $IoU(\%)$, ${{P}_{d}}(\%)$, ${{F}_{at}}(\%)$, and ${{F}_{a}}({10}^{-6})$. The best and the second-best results are highlighted in bold and underlined, respectively. \textit{EDGSP-Coa.} indicates the EDGSP under the coarse single-point prompt.}
\label{Tab1}
\centering
\setlength\tabcolsep{1.7mm}
\begin{tabular}{c|cccc|cccc|cccc|cccc}
\hline \hline
{\multirow{2}[1]{*}{Method}} & \multicolumn{4}{c|}{SIRST~\cite{ACM}} & \multicolumn{4}{c|}{NUDT-SIRST~\cite{DNA-Net}} & \multicolumn{4}{c|}{IRSTD-1k~\cite{Zhang_IS}} & \multicolumn{4}{c}{Mean}      \\  \cline{2-17} 
                        & $IoU\uparrow$    & ${{P}_{d}}\uparrow$     & ${{F}_{at}}\downarrow$ & ${{F}_{a}}\downarrow$   & $IoU\uparrow$    & ${{P}_{d}}\uparrow$     & ${{F}_{at}}\downarrow$ & ${{F}_{a}}\downarrow$   & $IoU\uparrow$   & ${{P}_{d}}\uparrow$    & ${{F}_{at}}\downarrow$ & ${{F}_{a}}\downarrow$   & $IoU\uparrow$   & ${{P}_{d}}\uparrow$    & ${{F}_{at}}\downarrow$ & ${{F}_{a}}\downarrow$   \\ \hline 

LESPS~\cite{Ying}      & 41.84  & 97.71  & 1.52  & 0.27 & 38.86  & 98.94  & 0.95  & 2.06 & 38.76 & 98.98 & 0.68  & 0.04 & 39.28 & 98.73 & 1.00  & 0.14 \\
COM~\cite{COM}        & 53.19  & 83.65  & 10.3   & 1.85  & 8.026   & 52.54  & 32.1    & 6.96 & 41.21 & 69.39 & 20.1  & 1.12 & 13.84 & 61.29 & 25.9 & 3.51 \\
MCLC~\cite{MCLC}      & 75.63  & 98.48  & 3.80  & 0.68 & 65.08  & 95.87  & 12.2 & 2.64 & 69.60 & 98.64 & 4.42  & 0.25 & 67.98 & 96.87 & 9.19  & 1.25 \\
MCGC~\cite{MCGC}      & 72.43  & 97.33 & 0.76  & 0.14 & 77.67  & 99.68  & 0.32  &  0.07  & 63.24 & \underline{99.66} & 0.34   &  \underline{0.02}  & 72.80  & 99.26 & 0.40  & 0.05 \\  \hline
SAM~\cite{SAM}       & 6.273     & \textbf{100.0}  &  0.76    &  0.14      &  8.466    &  \textbf{100.0}    & 0.74     &  0.16  & 2.428    &  \textbf{100.0}    &  1.02    &   0.06   &  6.41    &  \textbf{100.0}    & 0.80     &   0.11   \\
HQ-SAM~\cite{HQSAM}   &  6.657    &   \textbf{100.0}   &  \underline{0.38}    &   \underline{0.07}     & 8.734     &  \textbf{100.0}    &  \textbf{0}    &    \textbf{0}    & 2.426    &  \textbf{100.0}    &   \textbf{0}   &   \textbf{0}    &  6.61    &  \textbf{100.0}    &  \underline{0.07}    &  \underline{0.01}    \\ 
Robust SAM~\cite{RobustSAM}   & 78.89     & 98.86     &  \underline{0.38}    & \underline{0.07}       & 67.05     &  99.79    & 0.21     & \underline{0.05}       &  \underline{72.28}    &  99.32    & \underline{0.34}     &  \underline{0.02}     &  70.14    & 99.53     &  0.27    &   0.04    \\ 
EDGSP-Coa.    & \underline{83.05}  & \underline{99.62}  & \textbf{0}    & \textbf{0}    & \underline{94.23}  & \underline{99.89}  & \underline{0.11}  & 1.01 & 71.07 & \underline{99.66} & \underline{0.34}  & 0.47 & \underline{85.36} & \underline{99.80} &  0.13  & 0.62 \\ 
EDGSP     &  \textbf{83.83}   &  \textbf{100.0}     &  \textbf{0}     &  \textbf{0}     & \textbf{95.51}  &  \textbf{100.0}     & \textbf{0}     & \textbf{0}    & \textbf{73.80}  &  \textbf{100.0}   & \textbf{0}     & \textbf{0}    & \textbf{87.07} & \textbf{100.0}   & \textbf{0}     & \textbf{0}    \\ 
\hline
\end{tabular}
\end{table*}

\begin{figure*}[t]
    \centering
    \includegraphics[width=1\textwidth]{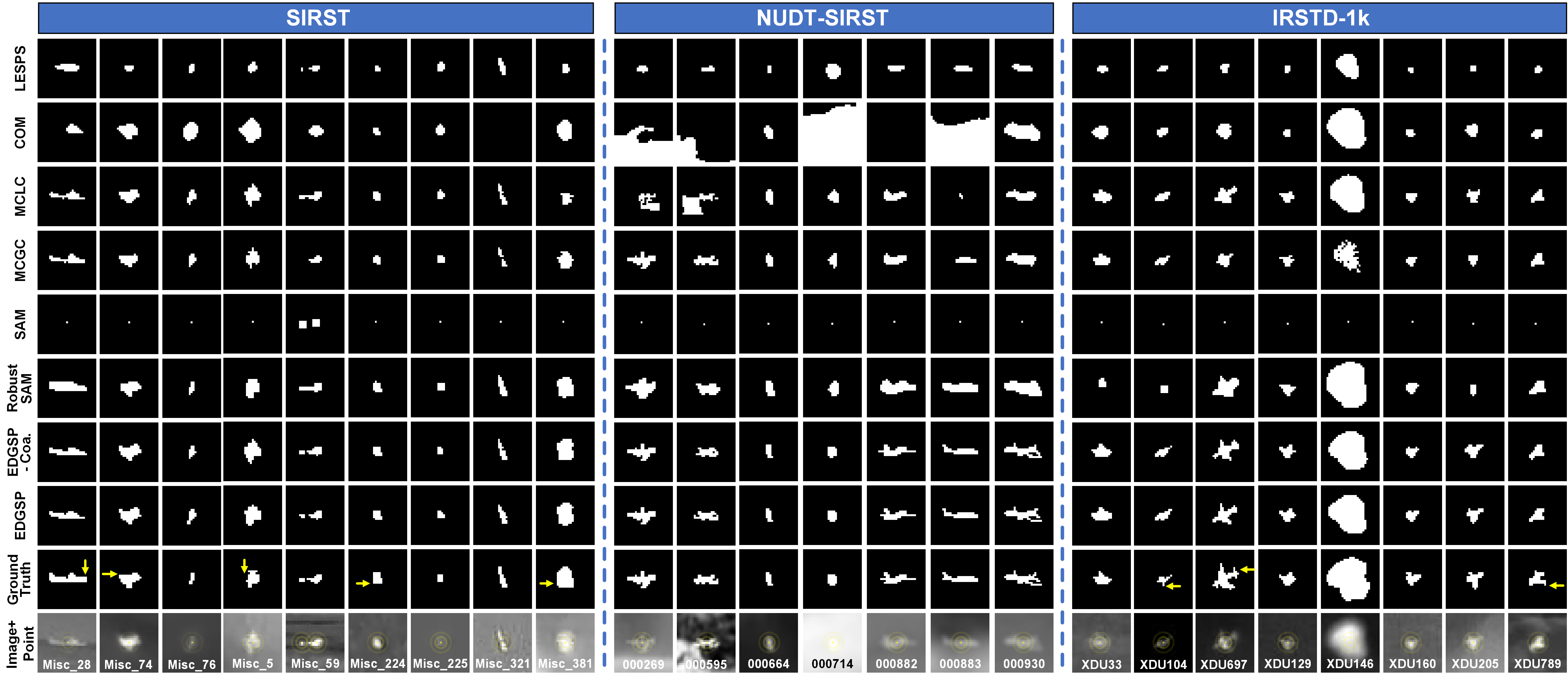}
    \caption{Visual results of seven IRSTLG methods on the SIRST, NUDT-SIRST, IRSTD-1k. \textit{EDGSP-Coa.} represents the pseudo label of EDGSP using the coarse single-point prompt. Small flaws in the ground truth are highlighted with arrows. For clarity, only annotated areas are shown.}
    \vspace{-2mm}
    \label{Fig5}
\end{figure*}

\subsection{Evaluation Metrics}
In line with previous works~\cite{MCLC},~\cite{MCGC},~\cite{COM}, a two-stage pipeline is presented to evaluate the effectiveness of pseudo labels and target detection results.
For target detection, we adopt three metrics: intersection over union (IoU), probability of detection (${{P}_{d}}$), and false-alarm rate (${{F}_{a}}$). 
Specifically, IoU serves as a pixel-level metric to assess the model's contour inference capability~\cite{ACM}.
Meanwhile, ${{P}_{d}}$ and ${{F}_{a}}$ are utilized as target-level evaluation metrics to assess localization capabilities, following the definitions in previous works, with a deviation threshold of 3 pixels for the target centroid~\cite{MTU},~\cite{DNA-Net},~\cite{SCTransNet}.
In the label generation task, we still use the three above-noted metrics, but the images to be evaluated are now represented by pseudo masks generated by the annotation algorithm.
Additionally, since the number of false alarms is more critical for the annotation task, 
we introduce a more refined target false-alarm rate (${{F}_{at}}$) for label generation, defined as follows:
\begin{equation}
{{F}_{at}}=\frac{T_{false}}{N_{all}},
\end{equation}
where the ${T}_{false}$ is false annotated target numbers and ${{N}_{all}}$ denotes all target numbers.

\subsection{Experimental Details} 
\subsubsection{Datasets}Following previous works~\cite{Ying}, our experiments are conducted on three mixed single-frame IRSTD datasets: SIRST~\cite{ACM}, NUDT-SIRST~\cite{DNA-Net}, and IRSTD-1k~\cite{Zhang_IS}, which include 427, 1327, and 1001 infrared images with one or more targets, respectively.
This means our method can annotate multiple datasets with just one set of weights.
The mixed dataset is split in a ratio of 8:4:1, with the fully labeled portion used for training EDGSP, the point prompt parts for generating pseudo labels to train the detection model, and the test set for evaluating detection performance.
In addition, we made necessary corrections to annotation errors in seven images from the NUDT-SIRST~\cite{DNA-Net} and IRSTD-1k~\cite{Zhang_IS} datasets, as shown in Fig.~\ref{Fig4}.

\begin{table*}[t]
\renewcommand{\arraystretch}{1.3}
\caption{Test results of five IRSTD methods after training with different point-based pseudo labels and full labels in terms of $IoU(\%)$, ${{P}_{d}}(\%)$, and ${{F}_{a}}({10}^{-6})$. The best and the second-best results are highlighted in bold and underlined, respectively.}
\label{Tab2}
\centering
\setlength\tabcolsep{1.2mm}
\begin{tabular}{l|c|c|c|c|c|c|c} \hline \hline
\multirow{2}{*}{Method}  & \multicolumn{6}{c|}{Point Generated Pseudo Label} & \multirow{2}{*}{Full Label}  \\  \cline{2-7} 
& COM~\cite{COM} & MCLC~\cite{MCLC}  & MCGC~\cite{MCGC} & Robust SAM~\cite{RobustSAM} & EDGSP-Coa.    & EDGSP   \\ \hline
ACM~\cite{ACM}    & 25.36/84.42/54.27        & 61.17/90.07/22.16  & 61.42/91.72/23.57     & 62.43/92.38/29.22   & 65.03/92.38/23.97   & 65.59/91.39/23.92     & 65.65/93.38/32.70  \\
RDIAN~\cite{RDIAN}   &19.48/75.17/78.26     & 64.94/94.04/28.54  & 68.38/93.38/16.96     & 65.57/93.38/24.43   & 73.27/93.05/21.26   & 74.68/93.71/18.23    & 72.64/92.72/13.93  \\
MTU-Net~\cite{MTU}  &  21.91/83.44/62.08    & 66.33/89.07/18.00  & 68.24/93.71/10.49     & 66.54/93.38/20.36   & 75.46/93.38/9.95    & 75.31/95.36/17.96    & 75.89/94.70/13.48  \\
DNA-Net~\cite{DNA-Net} & 22.61/79.14/49.81      & 66.12/93.05/15.29  & 67.80/93.38/6.60      & 67.91/94.37/12.48   & 81.52/94.70/7.55    & 81.53/96.69/8.14     & 81.63/95.70/9.00   \\
UIU-Net~\cite{UIUNet} & 28.73/80.13/40.70     & 66.07/93.05/44.06  & 69.55/93.71/4.26      & 66.57/93.70/6.02    & 81.14/94.70/8.73    & 82.70/96.03/6.02     & 82.46/96.68/7.46   \\ \hline
Average  & 23.62/80.46/57.02  & 64.93/91.86/25.61  & 67.08/93.18/12.38     & 65.80/93.44/18.50   & 75.28/93.64/14.29   & \textbf{75.96/94.64/14.85}    & \underline{75.65/94.64/15.31}  \\ \hline
\end{tabular}
\end{table*}

\begin{figure*}[t!]
    \centering
    \includegraphics[width=0.995\textwidth]{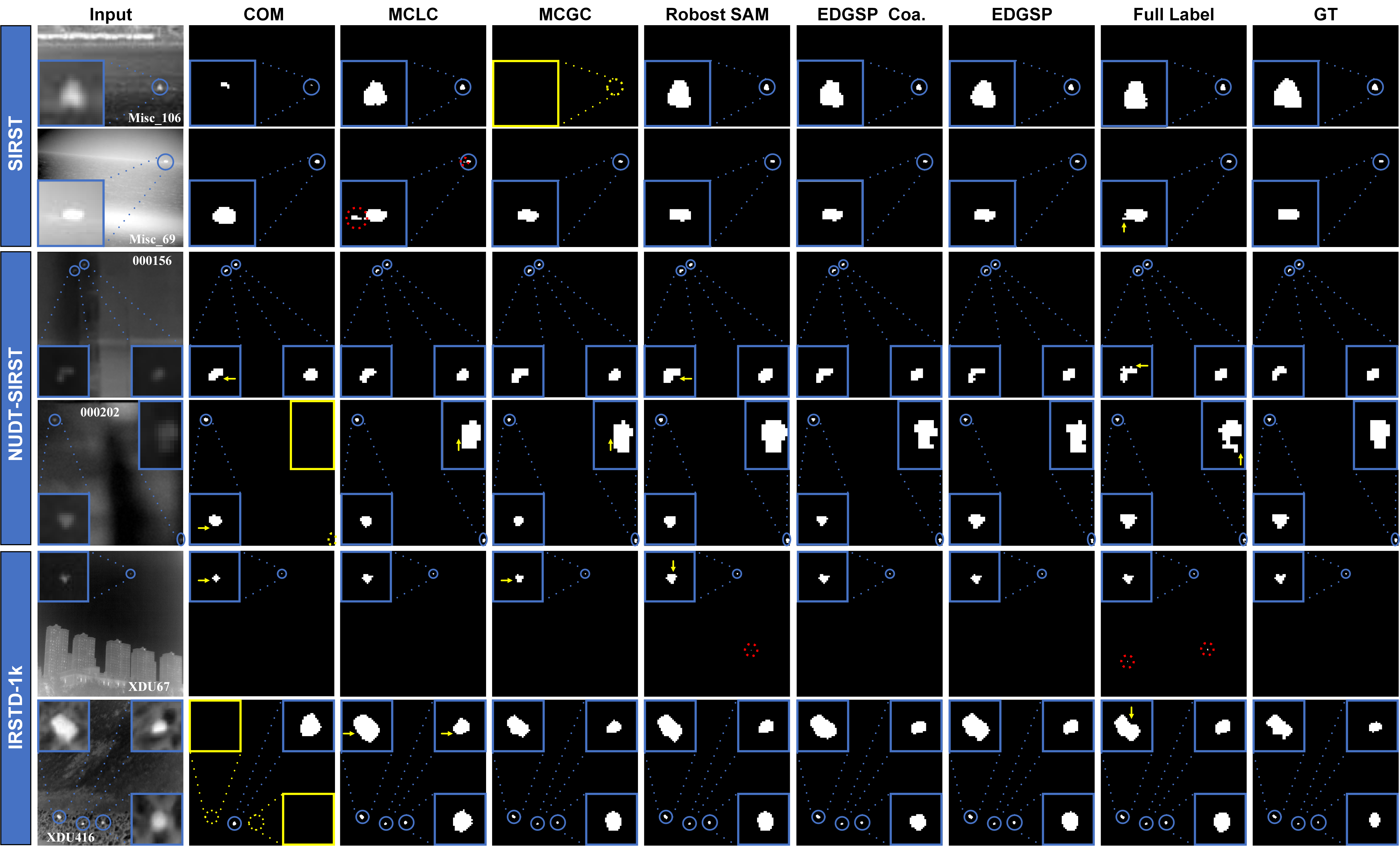}
    \caption{Detection results of DNA-Net trained with the different point prompt labels and the full label, respectively. Circles in blue, yellow, and red represent correctly detected targets, miss detections, and false alarms, respectively. The incomplete shapes are marked with arrows.
    }
    \vspace{-3mm}
    \label{Fig6}
\end{figure*}

\subsubsection{Implement Details}
We employ SCTransNet~\cite{SCTransNet} as the backbone of EDGSP, and the peak half-width of the Gaussian, i.e., $\sigma$ is set to 4. We will discuss them in Section~\ref{Tool} and Section~\ref{TEI}.
Following previous works~\cite{Ying},~\cite{UIUNet},~\cite{SCTransNet}, all training images are normalized and randomly cropped into 256 $\times$ 256 patches as inputs. 
To avoid over-fitting, random flipping and rotation are utilized to augment the training data. 
We employ the binary cross entropy~\cite{UIUNet} as the loss function to strengthen target boundaries. 
Subsequently, the Kaiming initialization method~\cite{KaiMing} is adapted to initialize the weights and bias for each layer.
All networks are trained from scratch and optimized by the Adam method~\cite{Adam}. After applying a Sigmoid function, the segmentation threshold of these models is set to 0.5.
The initial learning rate is 0.001 and is gradually decreased to $1\times{{10}^{-5}}$ using the cosine annealing~\cite{Cos}.
The batch size is 16. The training epochs are set as 1000 and 800 and we conduct early stop training after 800 and 600 epochs for label generation and target detection, respectively.
All experiments are implemented with PyTorch version 1.7.1 on a single Nvidia GeForce 3090 GPU, an Intel Core i7-12700KF CPU, and 32 GB of memory.

\subsubsection{Comparison Methods}
To evaluate the performance of pseudo mask, we compare EDGSP to the SOTA infrared small target label generation methods: MCLC~\cite{MCLC}, MCGC~\cite{MCGC} and COM~\cite{COM}, and learning-based single-point prompt annotation methods: SAM~\cite{SAM}, HQ-SAM~\cite{HQSAM} and Robust SAM~\cite{RobustSAM}. Furthermore, we also include the point-supervised IRSTD framework LESPS~\cite{Ying} with DNA-Net as the backbone, consistent with the original paper. 
For target detection, five high-performance IRSTD methods are used, including ACM~\cite{ACM}, RDIAN~\cite{RDIAN}, MTU-Net~\cite{MTU}, DNA-Net~\cite{DNA-Net}, and UIU-Net~\cite{UIUNet}.
Noticeably, \textit{these detection baselines are employed as a fair benchmark to evaluate the performance of pseudo masks generated by different label generation methods}.

\subsection{Label Generation Results}
\subsubsection{Quantitative Results}
As shown in Table~\ref{Tab1}, only EDGSP achieves ${{P}_{d}}$=100\%, ${{F}_{a}}$=0\%, ${{F}_{at}}$=0\% across three datasets at the same time. 
This indicates that our framework is currently the only one guaranteeing the most basic level of annotation quality.
Additionally, EDGSP's IoU surpasses the state-of-the-art (SOTA) labeling method (MCGC) by 14.27, demonstrating a superior shape prediction capability. 
Even with the coarse single-point prompt, EDGSP still outperforms other methods that use the centroid single-point prompt.
We also found that while the SAM and HQ-SAM exhibit strong localization capabilities, they have low IoU values and fail to infer reliable target profiles.
This observation indicates that vanilla SAM methods are unsuitable for small target annotation, given the significant differences between conventional general-purpose targets and IR small targets.

 
\subsubsection{Qualitative Results}
Fig.~\ref{Fig5} illustrates that, compared to other models, the pseudo masks generated by EDGSP more closely align with the target shape and better emphasize the profile features of the object.
Notably, masks such as Misc\_28, Misc\_74, 000595, XDU104, and XDU697 even exceed the quality of ground truth (GT). 
We also observe that COM and MCLC tend to merge the target with the background, while LESPS, MCGC, and Robust SAM struggle to accurately capture the morphology of infrared small targets. Additionally, SAM labels the target as a dot, making it difficult to provide contour information for small targets. 






\begin{table*}[t]
\renewcommand{\arraystretch}{1.3}
\caption{The improvement brought by the EDGSP to seven SOTA IRSTD methods on mixed datasets from SIRST, NUDT-SIRST, and IRSTD-1k in terms of $IoU(\%)$, ${{P}_{d}}(\%)$, ${{F}_{at}}(\%)$, ${{F}_{a}}({10}^{-6})$, $Flops(G)$, and $Params(M)$.}
\label{Tab3}
\centering
\setlength\tabcolsep{2.1mm}
\begin{tabular}{c|cccccc|cccccc} \hline \hline
\multirow{2}{*}{Method} & \multicolumn{6}{c|}{IRSTD Baselines}  & \multicolumn{6}{c}{IRSTD equipped with EDGSP} \\  \cline{2-13} 
         & $IoU\uparrow$  & ${{P}_{d}}\uparrow$ & ${{F}_{at}}\downarrow$ & ${{F}_{a}}\downarrow$  & ${Flops}\downarrow$  & ${Params}\downarrow$ & $IoU\uparrow$  & ${{P}_{d}}\uparrow$ & ${{F}_{at}}\downarrow$ & ${{F}_{a}}\downarrow$  & ${Flops}\downarrow$  & ${Params}\downarrow$     \\ \hline
ACM~\cite{ACM}        & 61.43   & 94.80  & 13.75  & 64.50  & 0.402 & 0.398  & 73.60 & 99.60           & 0.066      & 0.28       & 0.405    & 0.398  \\
ALCNet~\cite{ALCNet}     & 63.26   & 94.98  & 13.69  & 58.74  & 0.378 & 0.427  & 74.06 & 99.67           & 0.066      & 1.71       & 0.381    & 0.427 \\
MTU-Net~\cite{MTU}    & 71.62   & 94.15  & 12.16  & 28.31  & 6.194 & 8.221  & 84.20 & 100.0  & 0 & 0 & 6.215    & 8.221     \\
ISTDU~\cite{ISTDU}     & 80.12   & 96.74  & 9.103  & 32.18  & 7.944 & 2.752  & 84.07 & 99.93           & 0.066      & 1.11       & 7.973    & 2.752  \\
DNA-Net~\cite{DNA-Net}    & 80.53   & 96.90  & 5.249  & 15.28  & 14.26 & 4.697  & 86.78 & 100.0  & 0 & 0 & 14.28    & 4.697    \\
UIU-Net~\cite{UIUNet}    & 82.40   & 97.34  & 6.389  & 15.39  & 54.43 & 50.54  & 86.96 & 99.93           & 0 & 0 & 54.50    & 50.54  \\
SCTransNet~\cite{SCTransNet} & 83.43   & 97.28  & 5.781  & 8.622  & 10.12 & 11.19  & 87.07 & 100.0  & 0 & 0 & 10.16    & 11.19  \\ \hline
Average    & 73.81   & 95.79  & 9.993  & 34.32  & 13.39 & 11.18  & 82.19 & 99.85           & 0.041      & 0.43       & 13.42 &  11.18  \\
Difference & -       & -      & -      & -      &  -   &   -     & \textbf{ + 8.38} & \textbf{ + 4.06}           & \textbf{×0.35\%}    & \textbf{×1.25\%}    &  + 0.04   & \textbf{ + 0}     \\ \hline
\end{tabular}
\end{table*}

\subsection{Target Detection Results}
\label{Detection}
In this section, we use five IRSTD models as a fair benchmark, further evaluating the effectiveness of different pseudo labels for the detection task.

\subsubsection{Quantitative Results}
Table~\ref{Tab2} demonstrates that EDGSP outperforms other single-point prompt label generation methods on five IRSTD baselines. Although MCGC achieves the lowest ${{F}_{a}}$, it also has a low IoU score. 
Remarkably, EDGSP not only reduces time cost by 80\% but also outperforms full labels in detection tasks. 
Compared to the varying annotation styles of different annotators, deep labeling models offer greater accuracy and consistency. This is why the detection model trained with the proposed EDGSP outperforms those trained with manually annotated data.
Besides, we find that even with the coarse point prompt, EDGSP still achieves 99.5\% IoU values of pixel-wise labeling.

\subsubsection{Qualitative Results}
Fig.~\ref{Fig6} visualizes the detection results of DNA-Net on datasets with both point-label and full-label annotations.
The small targets to be detected vary in scale and contrast, and display significant differences in contours.
It can be observed that, whether using coarse or centroid annotations, the detection models trained with EDGSP outperform others in terms of false alarms, missed detections, and target contour accuracy, compared to full labels and other centroid annotation methods.
This reaffirms the applicability of the proposed methodology.


\subsection{Upgrade Tool: EDGSP}
\label{Tool}
In this part, we quantify the performance gains that point cues bring to representative IRSTD methods under the EDGSP framework.
As shown in Table~\ref{Tab3}, with the help of EGDSP, the average IoU and ${{P}_{d}}$ values for seven SOTA IRSTD methods, including ACM~\cite{ACM}, ALCNet~\cite{ALCNet}, MTU-Net~\cite{MTU}, ISTDU~\cite{ISTDU}, DNA-Net~\cite{DNA-Net}, UIU-Net~\cite{UIUNet} and SCTransNet~\cite{SCTransNet}, improved by 8.38\% and 4.06\%, respectively, while the ${{F}_{at}}$ decreased by a factor of 243.73. Flops and parameters remained nearly constant.
Three IRSTD methods (MTU-Net, DNA-Net, and SCTransNet) achieve accurate annotations. This indicates that EDGSP builds a reliable bridge between IRSTD and IRSTLG.
We also found a positive correlation between pseudo-label performance and detection results, indicating that future iterations of the IRSTD method can further enhance pseudo-label quality through EDGSP. Given that SCTransNet achieved the highest IoU score, we selected it as the backbone for EDGSP.

\begin{figure}[t]
    \centering
    \includegraphics[width=0.475\textwidth]{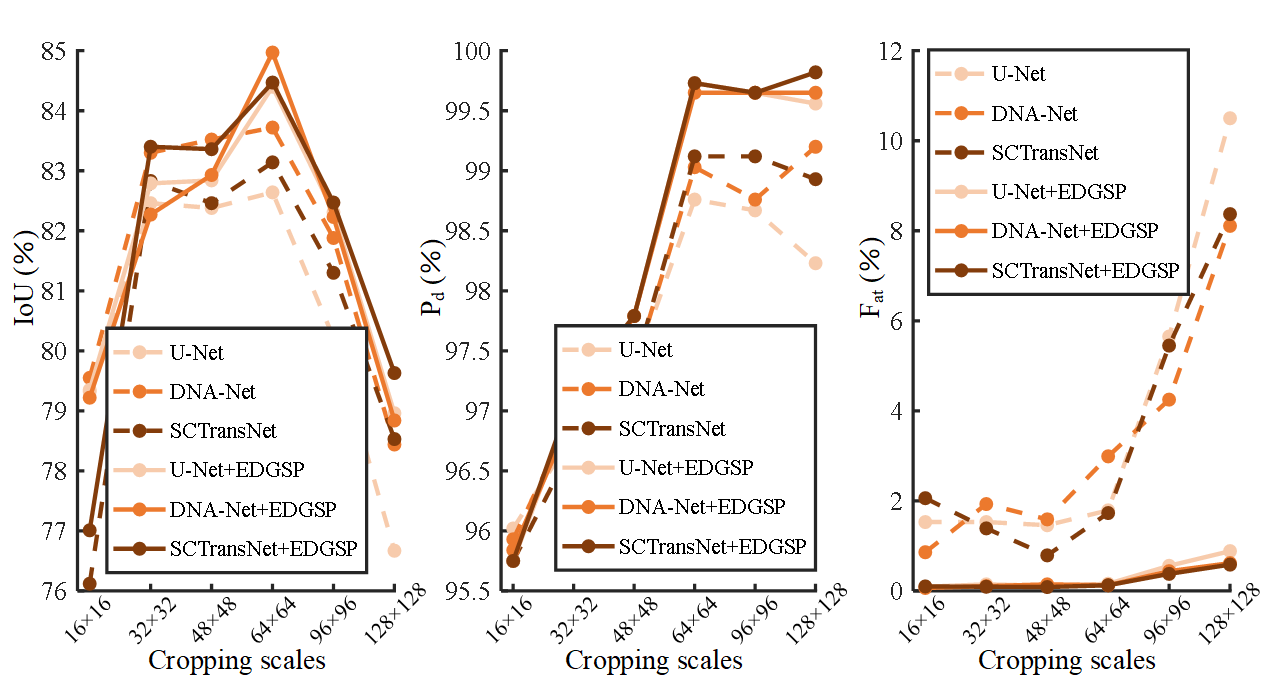}
    \caption{Quantitative results of three baselines at six cropping scales in terms of $IoU(\%)$, ${{P}_{d}}(\%)$, and ${{F}_{at}}(\%)$. Dashed and solid lines represent label generation results from direct center-cropping and from using EDGSP on the cropped image, respectively.
    }
    \label{Fig7}
\end{figure}

\subsection{Prompt-based Center-Cropping}
\label{Cropping}
As mentioned in Section~\ref{IIC}, cropping the region of interest based on the labeled position is a common way of target annotation.
Fig.~\ref{Fig7} showcases the results of three baselines (U-Net, DNA-Net, and SCTransNet) under prompt-based center-cropping.
With the crop size increasing, model performance initially improves and then declines, and fails to achieve high-performance labeling (${P}_{d}<100\%$, IoU$<$85).
This is because IR targets are multi-scale; small crop sizes fail to encompass large targets, while large crop sizes introduce excessive irrelevant padding at the image edges. 
Additionally, target overlap is challenging to manage when splicing the inferred image back into the original image.
In summary, missing boundary coordinate information diminishes the effectiveness of the formal center-cropping strategy. Therefore, our EDGSP directly processes the entire image.

\begin{table}[t]
\centering
\caption{Based on SCTransNet (SCT), ablation study of the single prompt embedding (SPE), TEI, rough edge map (REM), DPE, and  BBM on mixed datasets from SIRST, NUDT-SIRST, and IRSTD-1k. SPE means concatenating the single-point prompt with the original image as input.}
\label{Tab4}
\renewcommand{\arraystretch}{1.3}
\setlength\tabcolsep{1.3mm}
\begin{tabular}{c|cccccc|ccc}\hline\hline
Method &SCT     &SPE       & TEI        & REM    & DPE    & BBM     & $IoU\uparrow$ & ${{P}_{d}}\uparrow$ & ${{F}_{at}}\downarrow$  \\\hline
M1 & \ding{51}  &  \ding{55} &  \ding{55}     &  \ding{55}  &  \ding{55}  &  \ding{55}  & 83.43 & 97.28 & 5.78   \\\hline
M2 &\ding{51}  &\ding{51} &  \ding{55}     &  \ding{55}  &  \ding{55}  &  \ding{55}  & 86.16 & 99.67 & 1.73    \\\hline
M3 &\ding{51}  &\ding{51} & \ding{51}    &  \ding{55}  &  \ding{55}  &  \ding{55}  & 86.52 & 100.0 & 1.74   \\\hline
M4 &\ding{51}  &\ding{51} & \ding{51}    & \ding{51} &  \ding{55}  &  \ding{55}  & 86.74 & 99.93 & 1.93   \\\hline
M5 &\ding{51}  &  \ding{55} & \ding{51}    & \ding{51} & \ding{51} &  \ding{55}  & 87.02 & 100.0 & 1.59   \\\hline
Ours &\ding{51}  &  \ding{55} & \ding{51}    & \ding{51} & \ding{51} & \ding{51} & 87.07 & 100.0 & 0     \\\hline
\end{tabular}
\end{table}

\begin{figure}[t]
    \centering
    \includegraphics[width=0.475\textwidth]{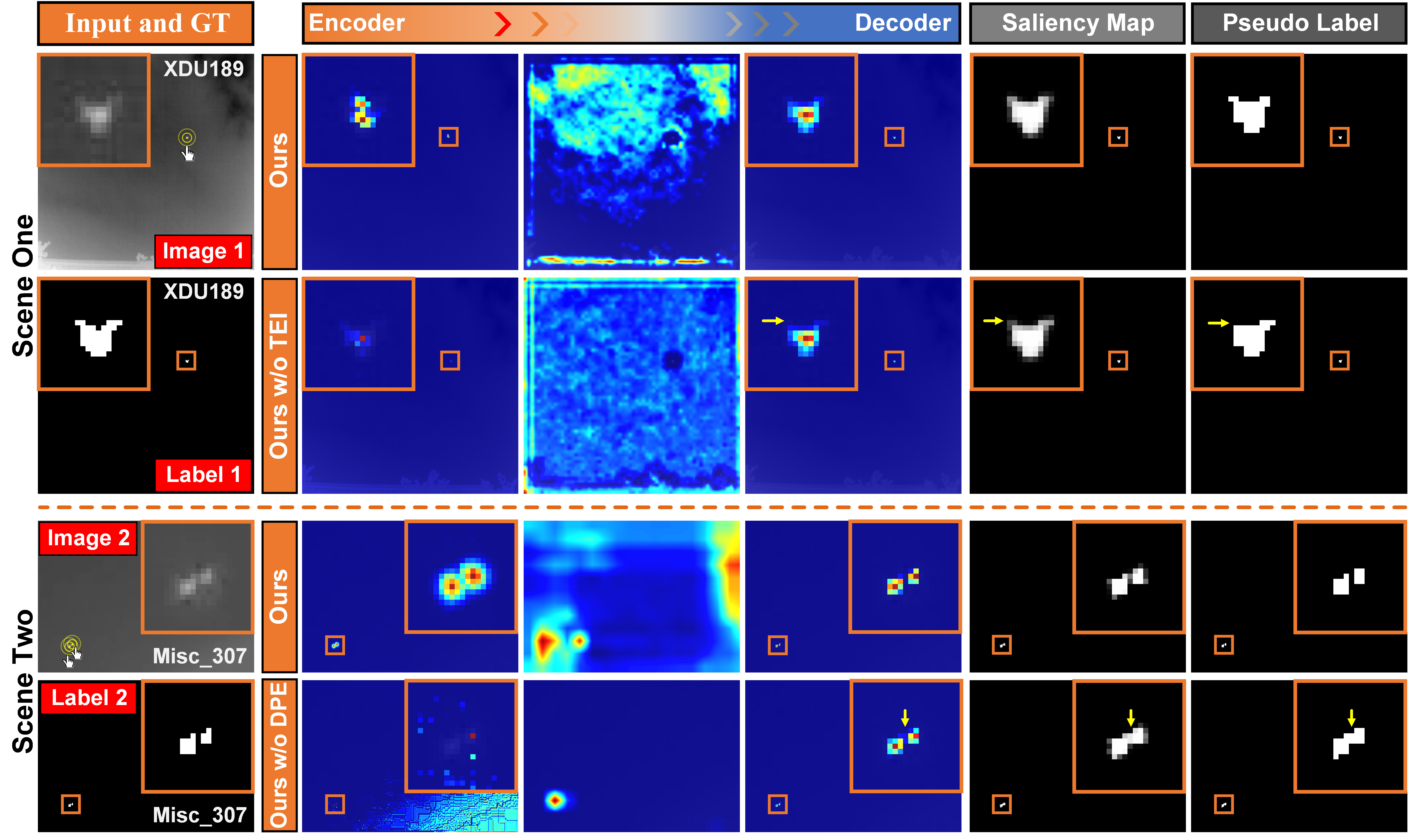}
    \caption{Visual maps of ablation study on TEI and DPE. Incomplete labeling and adhesion are highlighted with yellow arrows.}
    \label{Fig8}
\end{figure}

\begin{table}[t]
\renewcommand{\arraystretch}{1.3}
\caption{Ablation study of the embedding form on SIRST, NUDT-SIRST, and IRSTD-1k in terms of $IoU(\%)$, ${{P}_{d}}(\%)$, and ${{F}_{at}}(\%)$. The best results are boldfaced.}
\label{Tab5}
\centering
\setlength\tabcolsep{1.2mm}
\scalebox{0.95}{
\begin{tabular}{c|c|c|c} \hline \hline
 Method       &       Addition            &       Residual Attention           &      Channel Concatenation      \\ \hline 
U-Net            & 78.78/99.20/6.33          & 79.77/99.47/3.20                    & \textbf{80.89/99.60/1.67}        \\
DNA-Net                   & 86.04/99.80/2.79    & 86.43/99.87/1.07       & \textbf{86.56/99.87/0.87}                    \\
SCTransNet               & 86.03/99.74/3.19          & 86.36/99.93/0.93                    & \textbf{86.72/99.93/0.93}       \\ \hline                
\end{tabular}}
\end{table}

\begin{table*}[h]
\renewcommand{\arraystretch}{1.3}
\caption{Ablation study on the frequency and location of the embedding form on mixed SIRST, NUDT-SIRST, and IRSTD-1k datasets in terms of $IoU(\%)$, ${{P}_{d}}(\%)$, and ${{F}_{at}}(\%)$. The best results are highlighted in bold.}
\label{Tab6}
\centering
\setlength\tabcolsep{0.7mm}
\scalebox{0.86}{
\begin{tabular}{c|c|c|c|c|c|c|c|c|c} \hline \hline
\multirow{2}{*}{Method}  & \multicolumn{3}{c|}{Single Embedding}     & \multicolumn{3}{c|}{Double Embedding}    &  \multicolumn{3}{c}{Multiple Embedding} \\  \cline{2-10} 
            & Init         & Med        & Tail     & Init + Med   &  Init + Tail  & Med + Tail    &  Encoders  & Decoders   &  Encoders + Decoders  \\  \hline
U-Net       & 80.89/99.60/1.67 & 79.88/99.13/1.93 & 72.52/92.47/8.66     & 80.71/99.47/1.67  & 80.94/99.67/1.27    & 79.96/99.00/2.20   &80.59/99.67/1.80  &  79.76/94.27/11.93      & \textbf{81.10/99.73/1.87 }   \\
DNA-Net     & 86.56/99.87/0.87 & 85.34/99.60/2.26 & 82.34/99.27/6.45     & 89.91/99.80/1.60   & \textbf{87.15/99.93/1.47}   & 85.30/99.53/2.40   & 86.64/99.80/2.33 & 85.56/99.73/0.93   & 87.09/99.87/1.06     \\
SCTransNet & 86.72/99.93/0.93 & 85.20/99.30/2.80 & 82.02/97.60/4.33      & 86.89/99.86/2.26   & \textbf{86.45/100.0/1.59}  & 85.22/99.67/2.33   &  87.26/99.80/2.19  & 86.13/99.80/1.13 &  87.13/99.93/1.19  \\  \hline
\end{tabular}}
\end{table*}

\subsection{Ablation Study}
\label{ABL}
As shown in Table~\ref{Tab4}, SCTransNet is used as the baseline to validate the effectiveness of each proposed strategy.
In M2, we find that just incorporating single prompt embedding, the value of ${P}_{d} $ and IoU exceeds the full cropping strategies in Fig.~\ref{Fig7}, highlighting the need for whole image analysis.
With TEI, IoU further improved, and ${{P}_{d}}$ reached 100\%, which demonstrated it is capable of fleshing out target shapes.
After that, the incorporation of the rough edge map raised the IOU further but led to target sticking.
In M5, SPE is replaced with DPE. We can see that prompt embedding at the tail of the model highlights target edge differences and improves overall performance.
Finally, BBM accurately eliminates false alarms and EGDSP achieves accurate annotation. 

Observe the visual maps in Fig.~\ref{Fig8}. With the help of target energy initialization, the model quickly responds to the target area in the shallow layer and accurately predicts the left rotor of the drone. In scene two, without double prompt embedding~(\textit{Ours~w/o~DPE}), neighboring targets mix together.
We further observe that the attention map also responded in the background region, revealing that the local shape prediction remains connected to the global context modeling. 
Next, we will delve into a detailed discussion of the proposed DPE, TEI, and BBM.
Note that when discussing the DPE and  TEI, we do not apply BBM to process the generated labels. This ensures that the false alarms can be accurately observed.

\subsubsection{Double Prompt Embedding}
As a pioneer in learning-based IRSTLG work, we need systematically study the impact of point prompt embedding on annotation performance from three perspectives: form, frequency, and position, within general U-shaped structures.

\textit{a) Embedded Form:}
In this part, we consider three commonly used embedded forms: \textit{Addition}, \textit{Residual Attention}~\cite{CBAM}, and \textit{Channel Concatenation}~\cite{MIDeepSeg}.
Specifically, \textit{Addition} refers to the point prompt ${\mathbf{{I}_{p}}}$ being directly added to ${\mathbf{{I}_{i}}}$ as the model input; 
\textit{Residual Attention} means that we embed the point prompt ${\mathbf{{I}_{p}}}$ into a residual spatial attention module, which is utilized to modulate the first convolution layer of the detection backbone;
\textit{Channel Concatenation} denotes concatenating the point prompt ${\mathbf{{I}_{p}}}$ with the ${\mathbf{{I}_{i}}}$ among the channel-wise dimension as the model input.
Results are shown in Table~\ref{Tab5}, Among the three baselines (U-Net~\cite{U-Net}, DNA-Net~\cite{DNA-Net} and SCTransNet~\cite{SCTransNet}), 
\textit{Channel Concatenation} outperforms other strategies across the board.
It suggests that using the prompt as an independent semantic pattern can generate more accurate annotations. 
Thus, \textit{Channel Concatenation} is adopted in subsequent experiments.

\begin{table*}[t]
\centering
\renewcommand{\arraystretch}{1.3}
\caption{Comparison of different target energy initialization on SIRST, NUDT-SIRST, and IRSTD-1k in terms of $IoU(\%)$, ${{P}_{d}}(\%)$, and ${{F}_{at}}(\%)$. 
\textit{EDGSP w DNA.} and \textit{EDGSP w SCT.} refer to the DNA-Net and SCTransNet equipped with EDGSP, respectively.
$\xi$ denotes the attenuation coefficient in EACM, and $\delta$ represents the peak half-width of the Gaussian. The best results are highlighted in bold.}
\label{Tab7}
\setlength\tabcolsep{1.3mm}
\scalebox{1}{
\begin{tabular}{c|c|ccc|ccc}  \hline \hline
\multirow{2}{*}{Method} & \multirow{2}{*}{Non-initialization} & \multicolumn{3}{c|}{Euclidean Attenuation Coefficient Matrix~\cite{MCLC}} & \multicolumn{3}{c}{Gaussian Matrix} \\  \cline{3-8} 
                        &     & $\xi = 0.4$            & $\xi = 0.6$           & $\xi = 0.8$          & $\sigma =2$        & $\sigma =4$       & $\sigma =6$       \\ \hline
EDGSP w DNA.   & 86.69/99.93/0.87     & 86.80/99.87/1.53    & 87.11/99.87/1.27    &  86.59/99.93/1.33         &  86.74/100.0/0.93  & \textbf{86.78/100.0/0.73}       &   87.56/99.80/1.06      \\
EDGSP w SCT.     & 86.51/100.0/1.67   & 86.48/99.87/1.73   & 87.02/99.80/2.52    &   86.48/100.0/1.67          & 86.88/99.93/0.79  & \textbf{87.02/100.0/1.59}  & 87.81/99.87/2.73       \\ \hline
\end{tabular}}
\end{table*}

\begin{table}[t]
\caption{Based on different backbones with EDGSP, comparison of different false alarm elimination modules on SIRST, NUDT-SIRST, and IRSTD-1k in terms of $IoU(\%)$, ${{P}_{d}}(\%)$, and ${{F}_{at}}(\%)$. \textit{None} denotes the EDGSP without any post-processing module. The best results are boldfaced.}
\label{Tab8}
\centering
\renewcommand{\arraystretch}{1.3}
\setlength\tabcolsep{0.5mm}
\scalebox{0.86}{
\begin{tabular}{ccccc} \\ \hline \hline
Method     & None             &  TPM~\cite{DNA-Net}      & ERM~\cite{Ying}    & BBM         \\ \hline
\multicolumn{5}{c}{Centroid Single-point Prompt}                                           \\ \hline
EDGSP w U.       & 80.82/99.73/1.73  & 80.63/99.67/0     &  80.40/99.34/0.07  & \textbf{80.75/99.73/0.07} \\
EDGSP w DNA.    & 86.72/100.0/0.73  & \textbf{86.78/100.0/0}     &  86.45/99.73/0  & \textbf{86.78/100.0/0}  \\
EDGSP w SCT.  & 87.02/100.0/1.59  & \textbf{87.07/100.0/0}     &  86.91/99.87/0  & \textbf{87.07/100.0/0} \\ \hline
\multicolumn{5}{c}{Coarse Single-point Prompt}                                            \\ \hline
EDGSP w U.       & 80.66/99.53/1.47  & 70.61/94.27/0.13  & 81.23/98.73/0.47  & \textbf{81.64/99.47/0.47}   \\
EDGSP w DNA.    & 86.05/99.87/1.27  & 76.08/95.27/0    & 85.99/99.47/0.07 & \textbf{86.06/99.80/0.07}  \\
EDGSP w SCT.  & 85.48/99.87/3.00  & 75.57/95.14/0.13 & 85.22/99.27/0.13 & \textbf{85.36/99.80/0.13}  \\ \hline
\end{tabular}}
\end{table}


\textit{b) Embedded Frequency and Location:}
We divide the experiments into three groups to investigate the embedded frequency: single embedding, double embedding, and multiple embedding.
For single embedding, we explore three embedding locations:
\begin{itemize}
\item \textit{Init}: Using the prompt as the model input;
\item \textit{Med}: Embedding the downsampled prompt into the deepest layer of the U-shaped structure;
\item \textit{Tail}: The prompt is inserted at the end of the model.
\end{itemize}
For double embedding, we pair the above forms of single embedding in all possible combinations.
In terms of multiple embeddings, we embedded the prompt into all encoders, all decoders, and both encoders and decoders as input features, respectively.
Table~\ref{Tab6} reveals two conclusions:
1) Embedding at the input stage yields better annotation results, as it clearly indicates the high-resolution spatial location of the target. 
Delaying the embedding process may hinder the model from effectively focusing on key areas.
2) In the case of \textit{Init}, end-stage embedding significantly enhances ${{P}_{d}}$, suggesting that the model becomes more adept at distinguishing individual labeled instances during this phase.
Therefore, the \textit{Init + Tail} and the \textit{Encoders + Decoders} yield the best results.
Considering that BBM can eliminate false alarms, ensuring that every target is annotated becomes even more important.
Therefore, we adopt the \textit{Init + Tail} mode, dubbed double prompt embedding (DPE). It leverages early spatial cues and subsequent individual reinforcement to improve annotation performance.

\begin{figure}[t]
    \centering
    \includegraphics[width=0.490\textwidth]{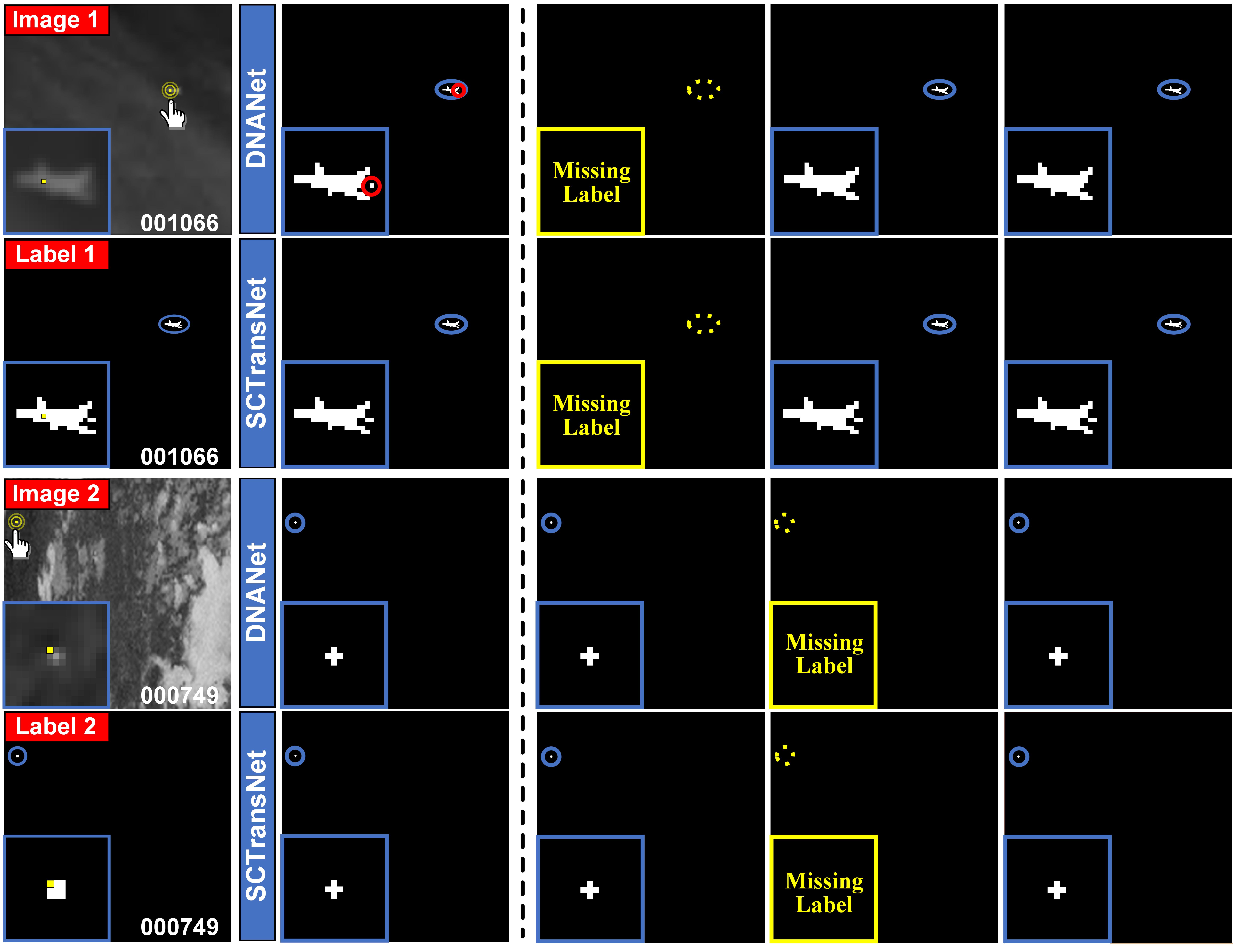}
    \caption{Visual results of different false alarm elimination modules with the coarse single-point annotation. False annotations are highlighted in the red circle. The missing labels are indicated with yellow circles.}
    \label{Fig9}
    \vspace{-3mm}
\end{figure}

\subsubsection{Target Energy Initialization}
\label{TEI}
In the previous work~\cite{MCGC}, the Euclidean attenuation coefficient matrix (EACM) is employed to extend the target form to highlight the correct region.
Table~\ref{Tab7} showcases the performance of EDGSP with two baselines (DNA-Net and SCTransNet) on two types of energy initialization with different hyperparameters. 
Specifically, $\xi$ denotes the attenuation coefficient in EACM (details in~\cite{MCGC}), and $\delta$ represents the peak half-width of the Gaussian in TEI.
We can find that using a Gaussian to initialize target energy is superior to EACM, which aligns with prior observations that infrared small targets exhibit a Gaussian-like energy distribution.
Therefore, the proposed EDGSP chooses a Gaussian with $\delta$ set to 4 to initialize the target distribution.


\subsubsection{Bounding Box-based Matching}
\label{BBMABL}
As mentioned in the manuscript, the bounding box-based matching (BBM) is designed as a false alarm elimination mechanism. This section introduces two comparison models: three-pixel matching (TPM) and eight-connective regions matching (ERM).
Specifically, TPM~\cite{DNA-Net} is defined as if the Euclidean distance between the centroid of a candidate target and the prompt coordinate is more than 3 pixels, the target is treated as a false alarm and removed. 
ERM~\cite{Ying} is defined as follows: if the prompt falls within the connective region of a target, that target is retained while other residual targets are discarded.
Table~\ref{Tab8} showcases the EDGSP under both centroid and coarse annotation modes across three post-processing strategies (TPM, ERM, and BBM).
In the case of centroid annotation, BBM and TPM achieved equally good results, while ERM mistakenly removed targets, leading to a decrease in ${{P}_{d}}$. 
Under the coarse annotation mode, the proposed BBM yielded the best results. Although TPM had the lowest false alarm rate, its ${{P}_{d}}$ dropped significantly. 
This is because the coarse prompt point deviated considerably from the target’s centroid, increasing the likelihood of incorrect removal. 
Fig.~\ref{Fig9} further illustrates two examples of incorrect removals by TPM and ERM under the coarse single-point prompt.

\begin{table}[t]
\caption{Efficiency of five label generation methods on images with different resolutions. 
${R_{256\times256}}$ and ${R_{512\times512}}$ represent the running times(s) of images with resolutions of 256 $\times$ 256 and 512 $\times$ 512, respectively.
The reported times are the averages of ten independent tests. The best and the second-best results are highlighted in bold and underlined, respectively.}
\label{Tab9}
\centering
\renewcommand{\arraystretch}{1.3}
\setlength\tabcolsep{2.6mm}
\scalebox{1}{
\begin{tabular}{l|ccc}  \hline \hline
Method              & IoU~$\uparrow$    & ${R_{256\times256}}\downarrow$ & ${R_{512\times512}}\downarrow$ \\  \hline
MCLC~\cite{MCLC}    & 67.98    & 0.2043   & 0.5099  \\  
MCGC~\cite{MCGC}    & \underline{72.80}  & 1.9987   & 4.7817 \\  
COM~\cite{COM}      & 13.84    & 1.5576   & 1.8805    \\  
Robust SAM~\cite{RobustSAM} & 70.14 & 0.2913      & \underline{0.3299}    \\ \hline
EDGSP                  &  \textbf{87.07}   & \textbf{0.0314}   &  \textbf{0.0346}   \\  
EDGSP (Only CPU)    &   \textbf{87.07} & \underline{0.1673}   &  0.4485      \\  \hline
\end{tabular}}
\end{table}

\subsection{Efficiency Analysis}
Table~\ref{Tab9} presents a comparison of the efficiency across five IRSTLG methods. 
The test images, with resolutions of 256 $\times$ 256 and 512 $\times$ 512, are sourced from the NUDT-SIRST~\cite{DNA-Net} and IRSTD-1k~\cite{Zhang_IS}, respectively. 
MCGC was implemented in MATLAB 2021a, while the other methods were executed in Pycharm 2020. Only Robust SAM and EDGSP utilized the GPU during testing, whereas the remaining methods were run on the CPU.
The result illustrates that the proposed EDGSP consistently outperforms the other methods in both annotation accuracy and real-time. For example, at the resolution of 256 $\times$ 256, the running time of EDGSP is 63.65 times faster than MCGC with the second-best IoU value. 
Importantly, even when running solely on the CPU, EDGSP achieves the best speeds compared to other methods under the same hardware conditions. This suggests that the proposed method can effectively meet daily annotation demands, even in low-power, resource-limited outdoor environments.

\section{CONCLUSION}
In this paper, we propose a novel annotation paradigm: enhancing the IRSTD network with the single-point prompt to accomplish high-precision IRSTLG.
Specifically, we introduce an EDGSP framework that includes target energy initialization for effective shape evolution, double prompt embedding to prevent label adhesion, and bounding box-based matching to eliminate false annotations.
Three backbones equipped with EDGSP achieve accurate annotation on three IRSTD datasets for the first time.
Excitingly, this is also the first time that a single-point generated mask outperforms full labels in practical IRSTD applications.
In the future, three promising directions may need to be further explored: 
1) Focus on coarse single-point prompts that match human markup habits; 
2) Leveraging SAM's capabilities to achieve robust performance with less infrared-labeled data; 
3) Developing a customized segmentation structure tailored for annotation models from scratch.

\bibliographystyle{IEEEtran}
\small\bibliography{IEEEabrv, reference}
\vspace{11pt}

 
 \begin{IEEEbiography}[{\includegraphics[width=1.1in, height=1.4in, clip, keepaspectratio]{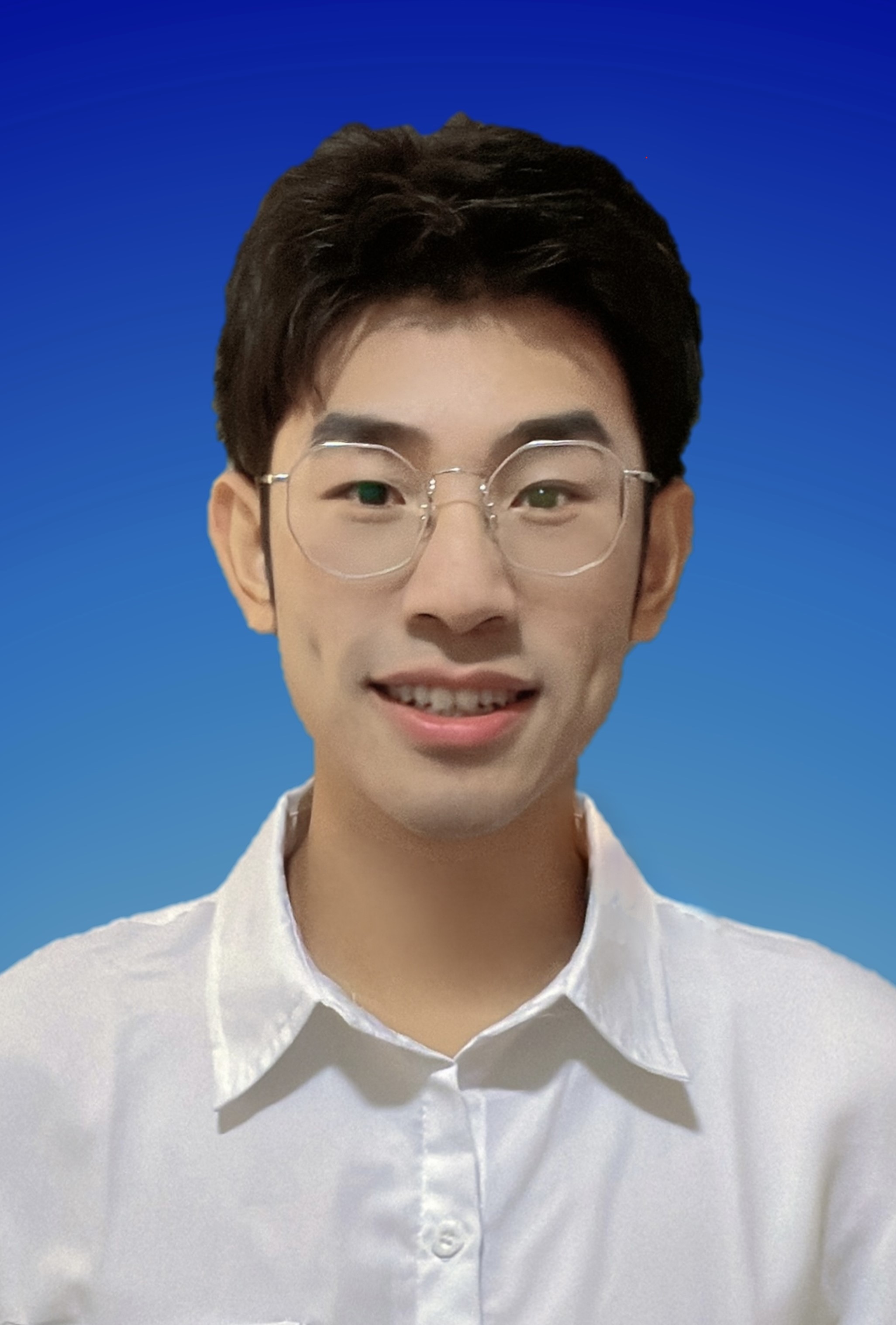}}]{Shuai Yuan}
 received the B.S. degree from Xi'an Technological University, Xi'an, China, in 2019. He is currently pursuing a Ph.D. degree at Xidian University, Xi’an, China. He is currently studying at the University of Melbourne as a visiting student, working closely with Dr. Naveed Akhtar. His research interests include infrared image understanding, remote sensing, and computer vision.
 \end{IEEEbiography}
 \vspace{-3mm}

 \begin{IEEEbiography}[{\includegraphics[width=1in, height=1.4in, clip, keepaspectratio]{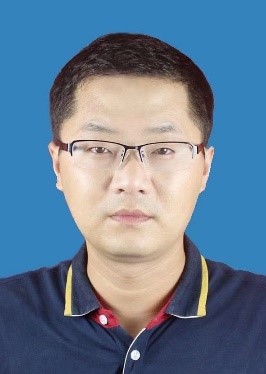}}]{Hanlin Qin}
received the B.S and Ph.D. degrees from Xidian University, Xi'an, China, in 2004 and 2010. He is currently a full professor at the School of Optoelectronic Engineering, Xidian University. He authored or co-authored more than 100 scientific articles. His research interests include electro-optical cognition, advanced intelligent computing, and autonomous collaboration.
\end{IEEEbiography}
 \vspace{-3mm}

\begin{IEEEbiography}[{\includegraphics[width=1in,height=1.25in,clip,keepaspectratio]{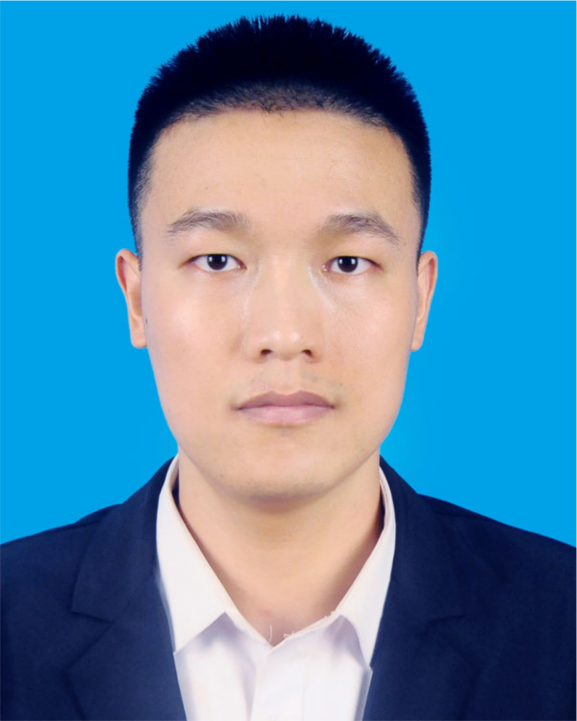}}]{Renke Kou} received the Ph.D. degree from Army Engineering University, Shijiazhuang, China, in 2024. He is currently working at the postdoctoral mobile station of the School of Aviation Engineering, Air Force Engineering University, Xi’an, China. His research interests include remote sensing, computer vision, and deep learning, with a focus on optimizing and deploying infrared small target detection algorithms to tackle real-world challenges. He has authored more than 10 peer-reviewed journal and conference papers such as TGRS (IEEE), JSTARS (IEEE), Remote Sensing (MDPI),  Sensors (MDPI), and Chinese Journal of Electronicsetc (IEEE), etc.
\end{IEEEbiography}
\vspace{-3mm}

 \begin{IEEEbiography}[{\includegraphics[width=1.5in, height=1.45in, clip, keepaspectratio]{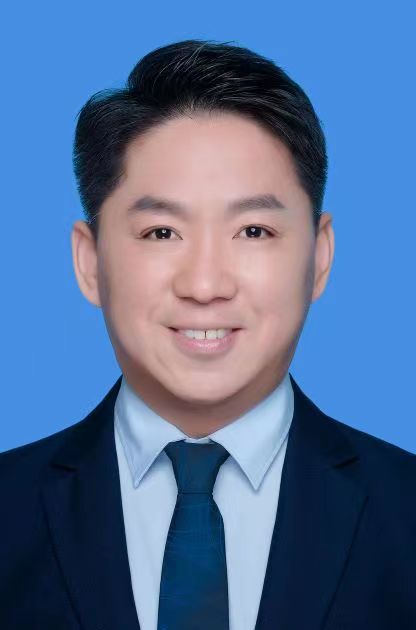}}]{Xiang Yan} received the B.S and Ph.D. degrees from Xidian University, Xi'an, China, in 2012 and 2018. He was a visiting Ph.D. Student with the School of Computer Science and Software Engineering, Perth, Australia, from 2016 to 2018, working closely with Prof. Ajmal Mian. He is currently an associate professor at Xidian University, Xi'an, China. His current research interests include image processing, computer vision and deep learning.
\end{IEEEbiography}
\vspace{-3mm}

\begin{IEEEbiography}[{\includegraphics[width=1.35in,height=1.4in, clip,keepaspectratio]{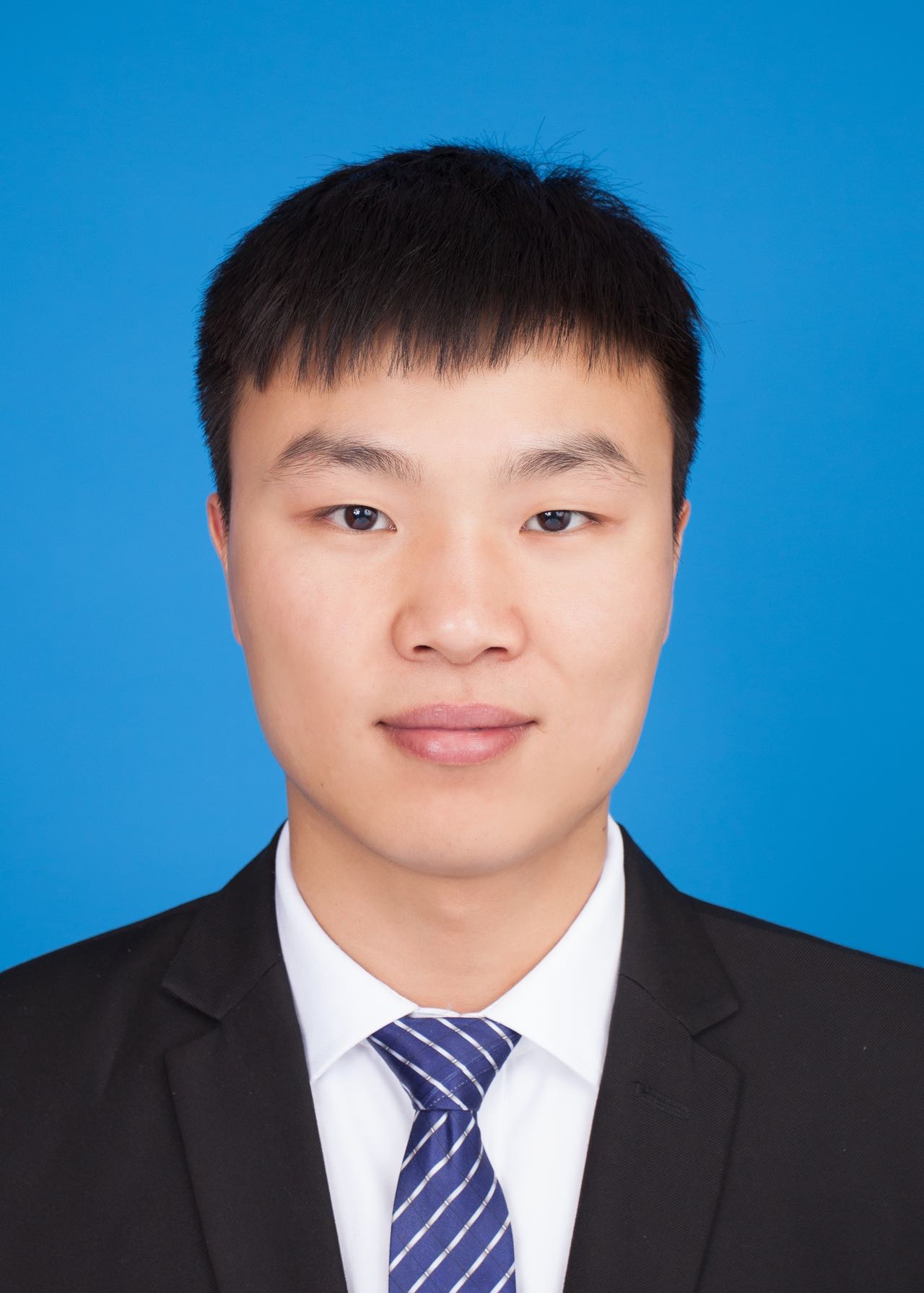}}]{Zechuan Li} 
received the M.S. degree in mechanical engineering from the Civil Aviation University of China (CAUC) in 2021. He is currently pursuing the Ph.D. degree with the College of Electrical and Information Engineering, Hunan University. He is currently studying at the University of Melbourne as a visiting student, working closely with Dr. Naveed Akhtar. His research interests lie in computer vision and point cloud processing. 
\end{IEEEbiography}
\vspace{-3mm}

\begin{IEEEbiography}[{\includegraphics[width=1.1in,height=1.3in, clip,keepaspectratio]{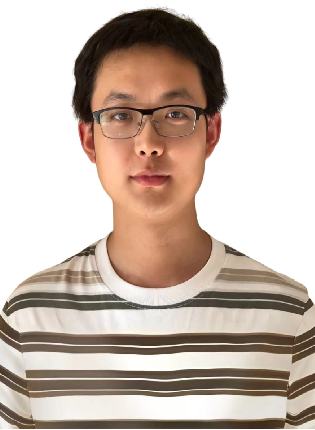}}]{Chenxu Peng}  received the M.S. at the College of Information Engineering, Zhejiang University of Technology. He is now working at Zhejiang SUPCON Information Co. Ltd. His current research interests include deep learning and medical image processing. 
He has won championships in multiple computer vision competitions organized by CCF, Kaggle, iFLYTEK, CVPR, PRCV, and ICPR.
\end{IEEEbiography}
\vspace{-3mm}

\begin{IEEEbiography}[{\includegraphics[width=1.1in,height=1.2in, clip,keepaspectratio]{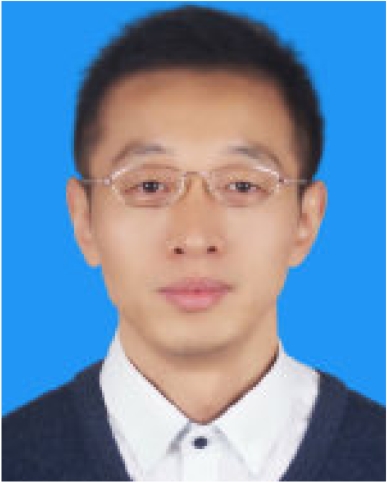}}]{Huixin Zhou}  (Member, IEEE) received the B.S.,
M.S., and Ph.D. degrees in optical engineering from
Xidian University, Xi’an, China, in 1996, 2002, and
2004, respectively.

He is currently the Vice-Dean and a Professor with the School of Physics, Xidian University. His research interests include remote sensing image processing, photoelectric imaging, real-time image processing, and target detection and tracking.

Dr. Zhou is also a Standing Member of the Optoelectronic Technology Professional Committee of the Chinese Society of Astronautics, a Senior Member of the Chinese Optical
Society, and a member of the Optical Society of America and the Shaanxi Provincial Physical Society. He is an Associate Editor of the journal of \textit{Infrared Physics} \& \textit{Technology}.
\end{IEEEbiography}
\vspace{-3mm}


\vfill

\end{document}